\documentclass[conference]{IEEEtran}
\IEEEoverridecommandlockouts
\usepackage{cite}
\usepackage{amsmath,amssymb,amsfonts}
\usepackage{algorithmic}
\usepackage{graphicx}
\usepackage{textcomp}
\usepackage{xcolor}
\usepackage{tikz}
\usepackage{array}
\usepackage{booktabs}
\usepackage{multirow}
\usepackage{url}
\usepackage{tabularx}
\usepackage{hyperref}
\usepackage{url}
\usepackage{float}
\usepackage{subcaption}
\usepackage{placeins}
\usetikzlibrary{shapes.callouts, positioning, arrows}
\def\BibTeX{{\rm B\kern-.05em{\sc i\kern-.025em b}\kern-.08em
    T\kern-.1667em\lower.7ex\hbox{E}\kern-.125emX}}
\begin{document}
\IEEEoverridecommandlockouts

\title{Transformers in Protein: A Survey\\
}

\author{\IEEEauthorblockN{Xiaowen Ling}
\IEEEauthorblockA{\textit{Department of Electronics and Computer Engineering} \\
\textit{Shenzhen MSU-BIT University}\\
Shenzhen, China \\
1120220517@smbu.edu.cn}
\and
\IEEEauthorblockN{Zhiqiang Li}
\IEEEauthorblockA{\textit{Department of Electronics and Computer Engineering} \\
\textit{Shenzhen MSU-BIT University}\\
Shenzhen, China \\
18046988607@163.com}
\and
\IEEEauthorblockN{Yanbin Wang\textsuperscript{\textasteriskcentered}}
\IEEEauthorblockA{\textit{Department of Electronics and Computer Engineering} \\
\textit{Shenzhen MSU-BIT University}\\
Shenzhen, China \\
wyb@smbu.edu.cn}
\and
\IEEEauthorblockN{Zhuhong You}
\IEEEauthorblockA{\textit{School of Computer Science and Technology} \\
\textit{Northwestern Polytechnical University}\\
Xi'an, China \\
zhuhongyou@nwpu.edu.cn}
}

\maketitle

\begingroup
\renewcommand\thefootnote{}\footnotetext{
\textsuperscript{\textasteriskcentered}Corresponding author: Yanbin Wang (email: wyb@smbu.edu.cn)
}
\endgroup

\begin{abstract}
As protein informatics advances rapidly, the demand for enhanced predictive accuracy, structural analysis, and functional understanding has intensified. Transformer models, as powerful deep learning architectures, have demonstrated unprecedented potential in addressing diverse challenges across protein research. However, a comprehensive review of Transformer applications in this field remains lacking. This paper bridges this gap by surveying over 100 studies, offering an in-depth analysis of practical implementations and research progress of Transformers in protein-related tasks. Our review systematically covers critical domains, including protein structure prediction, function prediction, protein-protein interaction analysis, functional annotation, and drug discovery/target identification.
To contextualize these advancements across various protein domains,  we adopt a domain-oriented classification system. We first introduce foundational concepts: the Transformer architecture and attention mechanisms, categorize Transformer variants tailored for protein science, and summarize essential protein knowledge. For each research domain, we outline its objectives and background, critically evaluate prior methods and their limitations, and highlight transformative contributions enabled by Transformer models. We also curate and summarize pivotal datasets and open-source code resources to facilitate reproducibility and benchmarking.
Finally, we discuss persistent challenges in applying Transformers to protein informatics and propose future research directions.  This review aims to provide a consolidated foundation for the synergistic integration of Transformer and protein informatics, fostering further innovation and expanded applications in the field.

\end{abstract}

\section{Introduction}
Protein informatics, a cornerstone of bioinformatics, is dedicated to deciphering protein structures, functions, and interactions. The exponential growth of biological data—particularly in proteomics—has intensified the demand for computational methods capable of efficiently and accurately interpreting complex protein datasets. While traditional approaches have laid a foundational understanding, their limitations in handling the scale and intricacy of modern protein data have spurred the adoption of more sophisticated models.

In this context, transformer models have emerged as a transformative solution across disciplines \cite{a1,a2,a3,liu2025pmanet,liu2025vul,sun2025ethereum}, owing to their ability to process variable-length sequences and model long-range dependencies via self-attention mechanisms. These attributes are especially valuable in protein informatics, where sequence-structure-function relationships often hinge on distal interactions and hierarchical patterns \cite{a4,a5}. Notably, transformer-based architectures (e.g., BERT, GPT) have achieved breakthroughs in protein structure prediction \cite{a6}, interaction analysis \cite{a7}, and functional annotation \cite{a8}.

\begin{figure*}[t]
    \centering
    \begin{subfigure}[b]{0.31\textwidth}
        \centering
        \includegraphics[height=0.15\textheight, width=\textwidth]{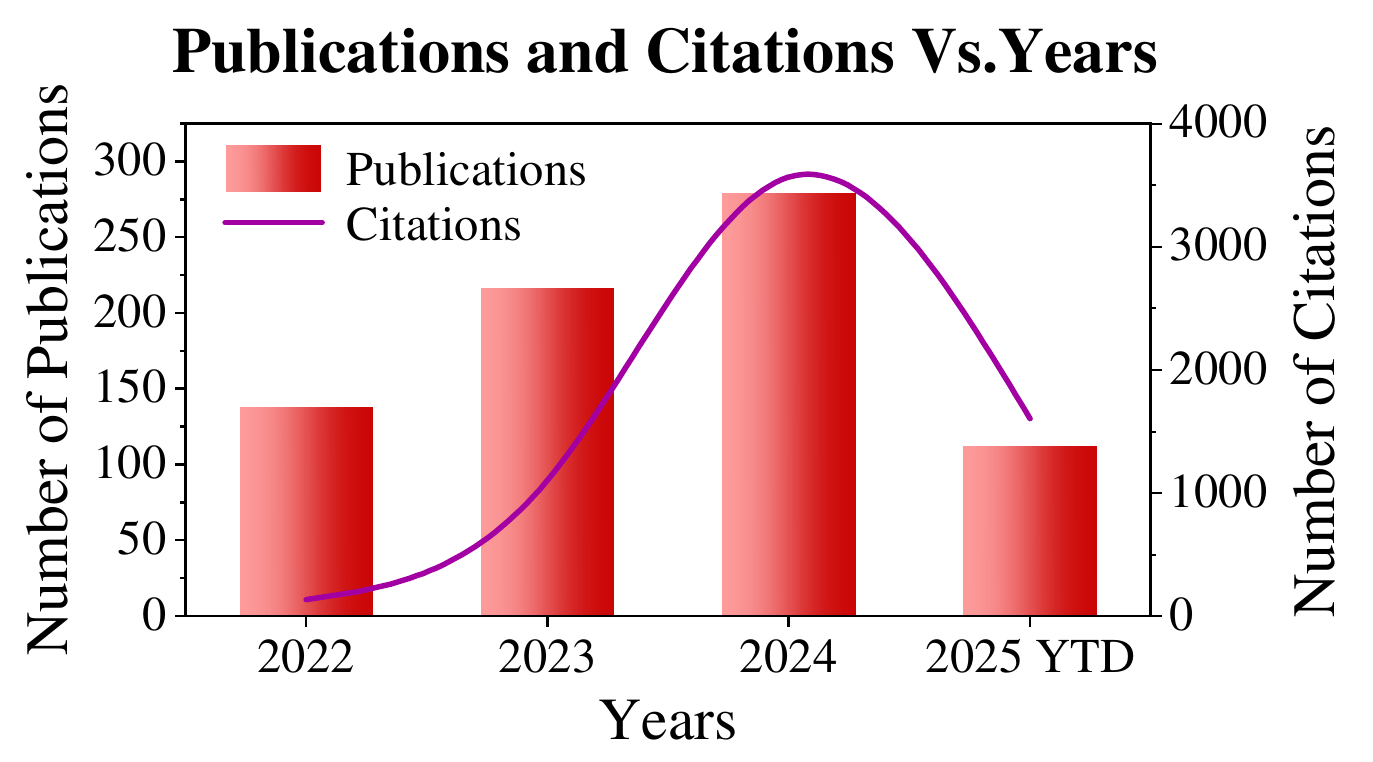}
        \caption{}
        \label{fig:sub1}
    \end{subfigure}
    \hspace{0.01\textwidth}
    \begin{subfigure}[b]{0.31\textwidth}
        \centering
        \includegraphics[height=0.15\textheight, width=\textwidth]{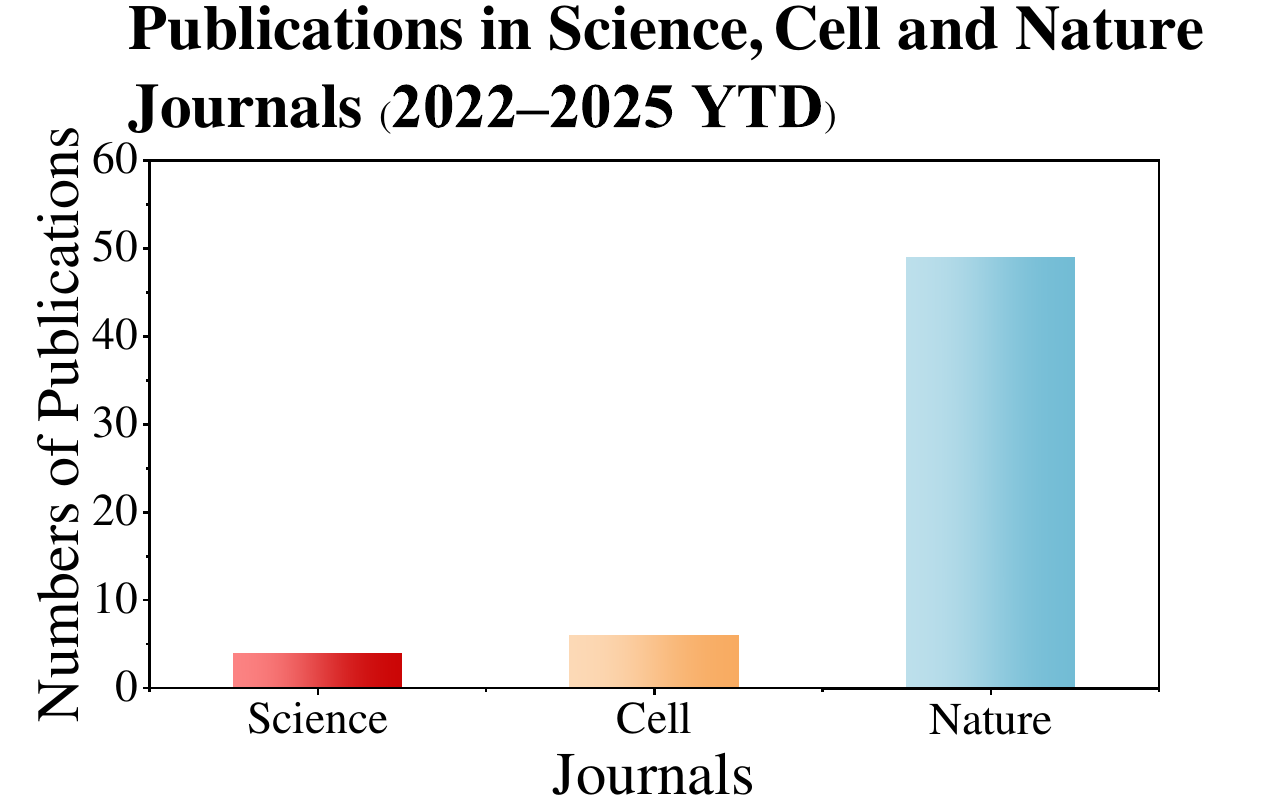}
        \caption{}
        \label{fig:sub2}
    \end{subfigure}
    \hspace{0.01\textwidth}
    \begin{subfigure}[b]{0.31\textwidth}
        \centering
        \includegraphics[height=0.15\textheight, width=\textwidth]{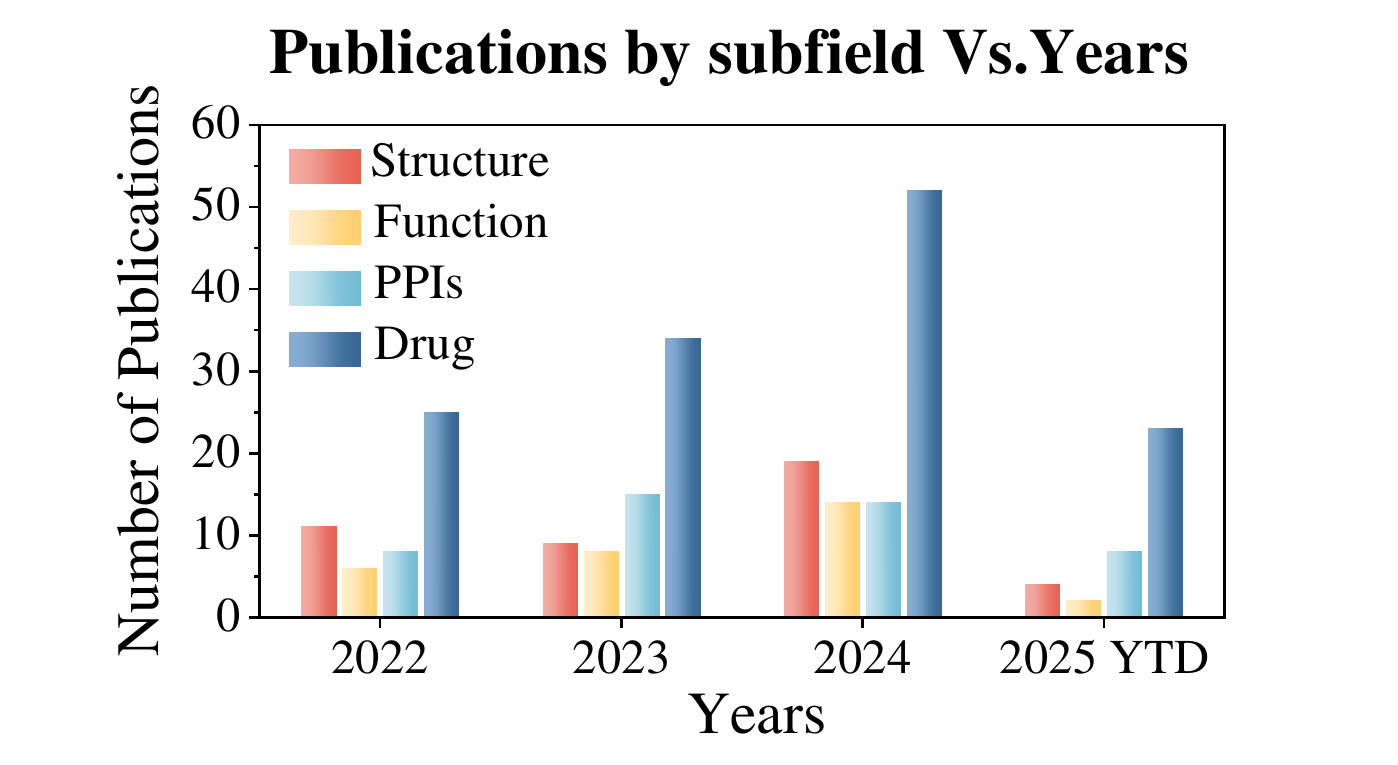}
        \caption{}
        \label{fig:sub3}
    \end{subfigure}
    \caption{Analysis of Transformer models in protein research using data from the Web of Science Core Collection. (a) Counts the number of publications and citations over the past four years, highlighting the growing influence of this research area. (b) Shows the distribution of publications across prestigious journals such as Nature, Science, and Cell. (c) Provides an overview of publications categorized by subfields from 2022 to 2025 YTD, illustrating the diverse applications of Transformer models in protein.}
    \label{fig:enter-label1}
\end{figure*}

The rising prominence of transformers in protein research is evident from publication trends. As depicted in Fig. \ref{fig:sub1}, the number of studies leveraging transformer models has surged in recent years, reflecting their growing adoption for protein-related tasks. To assess the impact of this trend in high-impact venues, we analyzed publications in Nature, Science, and Cell (including sub-journals) (Fig. \ref{fig:sub2}). Furthermore, we classified these works into key subfields of protein research (Fig. \ref{fig:sub3}), revealing the breadth of transformer applications in the discipline.

Transformer models have demonstrated remarkable potential in protein research, yet the field lacks a comprehensive synthesis of their applications, leaving critical gaps in understanding their full impact on protein informatics \cite{a9,a10}. This absence extends to systematic collections of essential resources, including benchmark datasets and algorithmic implementations that could accelerate progress. In response, our review undertakes a methodical examination of over 100 studies, analyzing both theoretical advances and practical applications across four pivotal domains: protein structure prediction, functional annotation, interaction analysis, and pharmaceutical discovery.

The review begins by establishing fundamental concepts, including transformer architectures, their protein-specific adaptations, and relevant biological principles. Building on this foundation, we employ a domain-oriented framework to evaluate each application area, first contextualizing its scientific objectives before critically assessing how transformer-based approaches compare to conventional methods \cite{a11,a12,a13}. Our analysis not only highlights transformative contributions but also curates vital computational resources, compiling authoritative datasets and state-of-the-art codebases to facilitate reproducibility and future innovation.

While documenting these advances, we identify persistent challenges in deploying transformers for protein studies, particularly regarding computational efficiency, data requirements, and model interpretability \cite{a14,a15}. These limitations inform our proposed research directions, which emphasize architectural specialization for biological tasks and hybrid approaches combining machine learning with biophysical principles. By synthesizing current knowledge and outlining pathways for improvement, this work seeks to strengthen the synergy between artificial intelligence and protein science, ultimately enabling novel methodologies that could reshape both fields \cite{a16}.

\section{Foundations}
Originally developed for NLP, Transformer models have proven highly effective in protein informatics due to their ability to process sequential data and capture long-range dependencies in protein sequences. Their success stems from two key innovations: (1) the self-attention mechanism, which overcomes traditional recurrent architectures’ limitations in modeling distant sequence relationships, and (2) the pre-training paradigm using large-scale (un)labeled data with (self-)supervised learning, followed by task-specific fine-tuning \cite{a2,a17,a18}.

The paper is organized as follows: Sections \ref{SecA} and \ref{SecB} introduce Transformer fundamentals, Section \ref{SecC} discusses protein-specific variants, and Section \ref{SecD} reviews essential protein biology. This foundation supports our analysis of Transformer applications in protein informatics (Section \ref{SecIII}).

\subsection{Transformer Architecture and Self-Attention Mechanism}\label{SecA}

The Transformer model, a novel network architecture introduced by \cite{a1}, entirely discards recurrence and convolution, relying solely on attention mechanisms. This design allows the model to dynamically assess the importance of different sequence elements. Their paper highlights that experiments on two machine translation tasks demonstrate the model's superior quality, greater feasibility, and significantly reduced training time. This self-attention mechanism computes the relationships between all pairs of input elements simultaneously, enabling efficient parallel processing, which is crucial for handling large datasets in protein informatics \cite{a3} \cite{a19}.

Self-attention dynamically models pairwise relevance between elements in a sequence (e.g., quantifying co-occurrence probabilities between words in a sentence) to explicitly capture intrasequence dependencies. As a core computational primitive in Transformer architectures, this mechanism enables differentiable relational reasoning through full-sequence interaction modeling for structured prediction tasks. Computationally, each self-attention layer iteratively refines the feature representation at every sequence position via global contextual aggregation.
This is achieved by defining three learnable weight matrices:
\begin{itemize}
    \item Query matrix: $W^Q \in \mathbb{R}^{d \times d_q}$
    \item Key matrix: $W^K \in \mathbb{R}^{d \times d_k}$ 
    \item Value matrix: $W^V \in \mathbb{R}^{d \times d_v}$
\end{itemize}
where $d_q = d_k$. The input sequence $X$ is projected as:
\begin{align*}
    Q &= XW^Q \\
    K &= XW^K \\
    V &= XW^V
\end{align*}

% 自注意力公式
The output $Z \in \mathbb{R}^{n \times d_v}$ of the self-attention layer is computed by:
\begin{equation}
    Z = \text{softmax}\left( \frac{QK^T}{\sqrt{d_q}} \right)V
\end{equation}

\begin{figure*}
\centering
    \includegraphics[width=\textwidth]{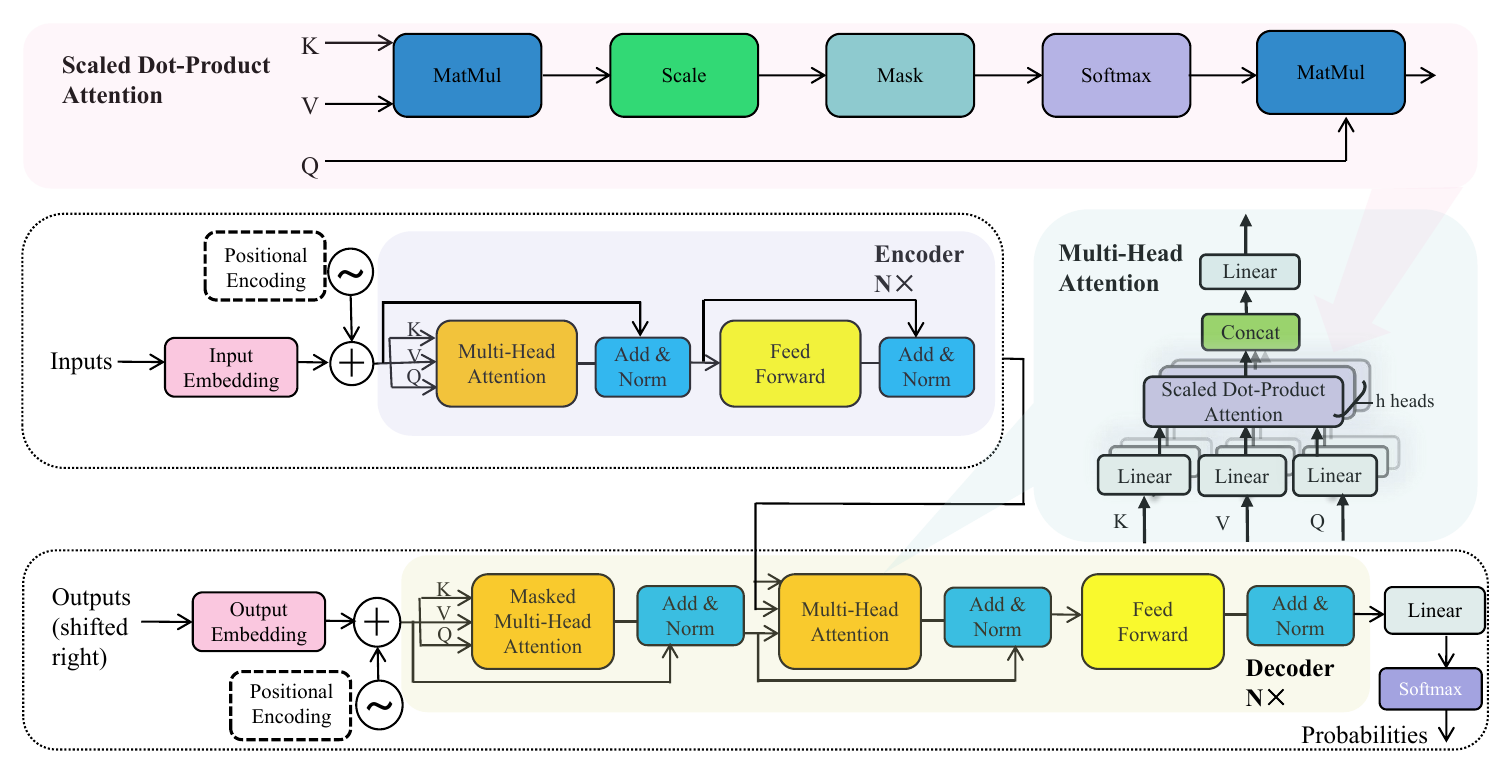}
    \vspace{-10pt} %
    \caption{Architecture of the Transformer Model\cite{a1}Originally proposed for machine translation tasks, the Transformer model transforms a source-language input sequence into a target-language output sequence through a dual-path architecture: (1) The encoder pathway processes input tokens via embedding projection and N identical blocks containing multi-head attention and feed-forward layers to generate continuous representations, while (2) the decoder pathway autoregressively produces output tokens by jointly attending to both the encoded source sequence and its own right-shifted inputs - where target sequences are shifted rightward with prepended ⟨SOS⟩ tokens during training to prevent trivial copying, while the loss is computed against the original sequence appended with ⟨EOS⟩ tokens. Both encoder and decoder stacks employ N modularized layers integrating multi-head attention mechanisms, position-wise feed-forward networks, and residual connections with layer normalization, enabling effective modeling of cross-lingual structural dependencies.}
    \label{fig:enter-label}
\end{figure*}
For any given entity within the sequence, the self-attention mechanism fundamentally operates by:

1.Computing the dot products between its query vector and all key vectors in the sequence,

2.Applying softmax normalization to these dot products to obtain attention weights.

The updated representation of each entity is subsequently computed as a convex combination (weighted sum) of all sequence entities, where the combination coefficients correspond to the derived attention weights (Fig.\ref{fig:enter-label}, the third row-left block).

\textbf{Masked Self-Attention Mechanism:} In standard self-attention layers, each position can attend to all other positions in the sequence. However, for autoregressive sequence generation tasks, the decoder employs masked self-attention to prevent information leakage from future positions. This is implemented through an element-wise masking operation:
\begin{equation}
    \text{Attention}_{\text{masked}}(\mathbf{Q},\mathbf{K},\mathbf{V}) = \text{softmax}\left( \frac{\mathbf{Q}\mathbf{K}^\top}{\sqrt{d_k}} \odot \mathbf{M} \right)\mathbf{V}
\end{equation}

where:
\begin{itemize}
    \item $\mathbf{M} \in \{0,-\infty\}^{n \times n}$ is an upper triangular mask matrix:
    \begin{equation}
    M_{ij} = \begin{cases}
        0 & \text{if } i \geq j \\
        -\infty & \text{if } i < j
    \end{cases}
\end{equation}
    \item $\odot$ denotes the Hadamard (element-wise) product
    \item $d_k$ is the dimension of key vectors
\end{itemize}

The masking operation ensures that when predicting the $i$-th position:
\begin{itemize}
    \item Only positions $j \leq i$ contribute to the attention weights
    \item Future positions ($j > i$) receive attention scores of $0$ (after softmax)
\end{itemize}

This implementation maintains the parallel computation advantages of self-attention while enforcing the autoregressive property required for sequence generation.

\textbf{Multi-Head Attention Mechanism:} The mechanism employs $h$ parallel self-attention heads ($h\!=\!8$ in \cite{a1}), each with independent weight matrices $\{\mathbf{W}^Q_i, \mathbf{W}^K_i, \mathbf{W}^V_i\}$ for $i \in [0,h\!-\!1]$. Given input $\mathbf{X}$, it concatenates all heads' outputs $[\mathbf{Z}_0, ..., \mathbf{Z}_{h-1}] \in \mathbb{R}^{n \times h d_v}$ and projects them through $\mathbf{W} \in \mathbb{R}^{h d_v \times d}$ (Fig.\ref{fig:enter-label}, the third row).

The fundamental distinction between self-attention and convolutional operations lies in their filter generation mechanism. Unlike convolution's static filters that remain fixed regardless of input, self-attention dynamically computes input-dependent filters. This approach exhibits two key advantages: (1) permutation invariance, making it robust to input ordering and variable numbers of input points, and (2) the ability to process irregular, non-grid structured data natively.

Research has demonstrated that when augmented with positional encodings, self-attention can theoretically subsume convolutional operations as a special case for local feature extraction\cite{a170}. Subsequent empirical studies by Cordonnier et al.\cite{a171}confirmed that properly configured multi-head self-attention can replicate and extend convolutional behaviors. Specifically, self-attention offers superior expressiveness by simultaneously learning both global and local features while adaptively determining optimal kernel weights and receptive fields - capabilities that parallel advanced techniques like deformable convolutions\cite{a172}.

\subsubsection{Single-head vs. Multi-head Self-Attention in Protein Transformers}

In the context of protein modeling, self-attention mechanisms serve as the core component for capturing relationships between amino acid residues. While single-head self-attention applies a single attention distribution across the sequence, its representational capacity is limited. In contrast, modern protein-related Transformer models universally adopt multi-head self-attention, which enables the model to attend to multiple subspaces simultaneously. This is particularly beneficial in protein structure prediction and sequence modeling tasks, where both local motifs and long-range interactions are crucial. For example, models like AlphaFold, ESM-Fold, and OmegaFold rely on multi-head attention to effectively capture diverse structural and evolutionary dependencies. As such, multi-head self-attention has become a standard design choice in protein Transformer architectures.

\begin{figure*}[h!]
    \centering
    \includegraphics[width=\textwidth]{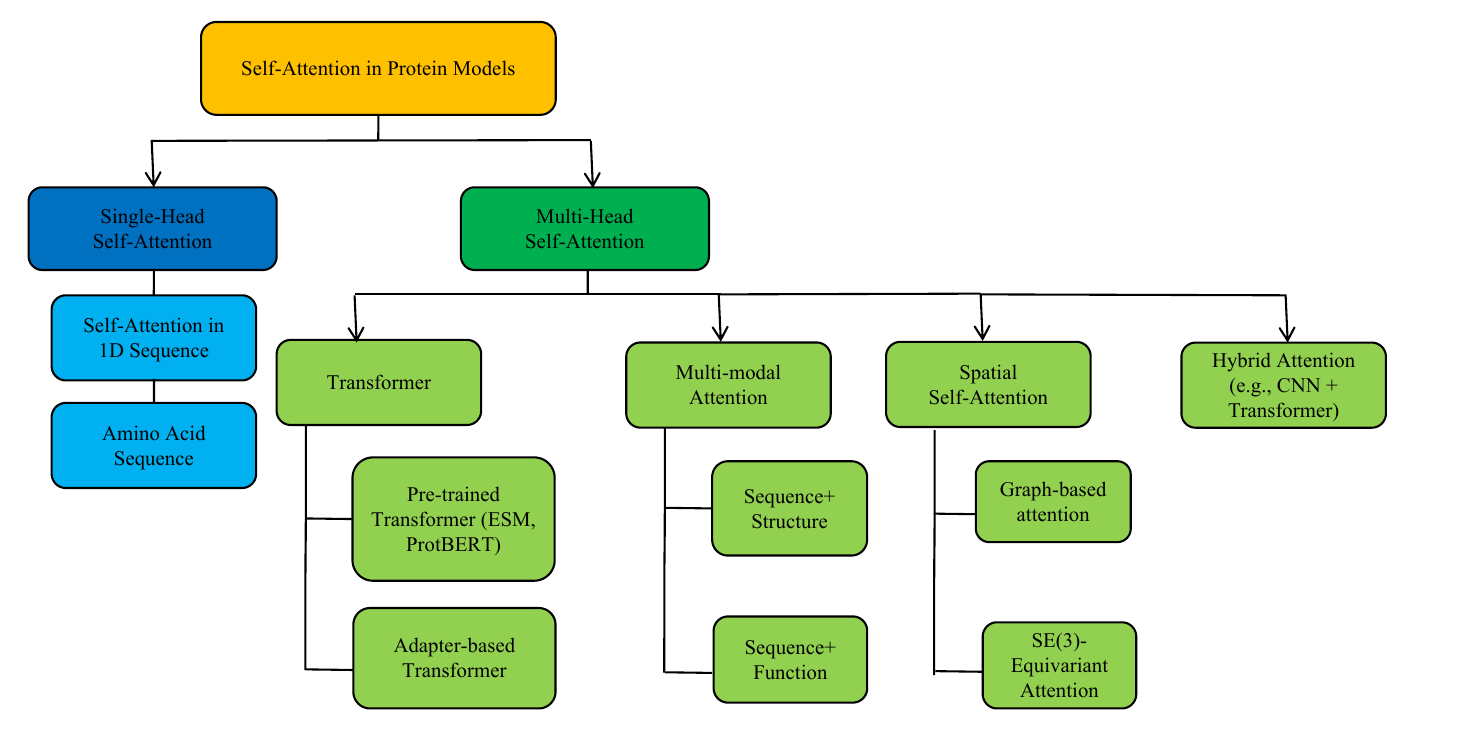} % 使用正确的文件名
    \caption{Hierarchical Classification of Self-Attention Mechanisms in Protein Models. This diagram illustrates the categorization of self-attention mechanisms adapted from vision transformer literature. The top-level distinction is made between Single-Head Self-Attention and Multi-Head Self-Attention. Under Multi-Head Self-Attention, several extensions are shown, including standard Transformer models, Pre-trained Transformers such as ESM and ProtBERT, Adapter-based Transformers, and Multi-modal Attention mechanisms that integrate sequence and structure or sequence and function. Additional branches include Spatial Self-Attention, which encompasses Graph-based attention and SE(3)-Equivariant Attention, and Hybrid Attention, which combines CNN with Transformer architectures.}
    \label{fig:protein_attention}
\end{figure*}

\subsection{(Self-)supervised pre-training}\label{SecB}
Transformer models applied to protein-related tasks typically adopt a two-stage training paradigm, inspired by advances in natural language processing and adapted to the biological sequence domain. In the first stage, models are pre-trained on large-scale protein datasets, either in a supervised or self-supervised manner, to learn generalizable representations of protein sequences or structures. In the second stage, the pre-trained models are fine-tuned on downstream tasks such as protein function prediction, structure modeling, subcellular localization, and protein–protein interaction prediction.

Due to the scarcity and high cost of experimental annotations in the biological sciences, self-supervised learning (SSL) has emerged as a particularly effective approach for pre-training. SSL enables models to learn meaningful representations from unlabeled protein data by solving carefully designed proxy tasks. One widely adopted strategy is masked language modeling, inspired by BERT, where a proportion of amino acid residues in protein sequences are masked and the model is trained to recover them. This formulation has been used in models such as ProtBERT\cite{a173}, ESM\cite{a174}, and PeptideBERT\cite{a175}, effectively capturing both local and global dependencies within sequences.

Other SSL objectives in protein modeling include inter-residue distance prediction\cite{a176}, contact map reconstruction, and learning from multiple sequence alignments (MSAs) to incorporate evolutionary relationships\cite{a177}. Recently, cross-modal SSL strategies have also been explored, where models jointly process sequence data and structural inputs (e.g., 3D coordinates or contact matrices), enabling them to learn biologically grounded representations\cite{a178}.

Broadly, current SSL approaches in protein modeling can be categorized into three main types:
\begin{enumerate}
    \item Generative methods, which aim to reconstruct protein sequences or structures from partial or noisy inputs. Examples include masked token prediction, sequence inpainting, and structure reconstruction from sequence embeddings.
    \item Context-based methods, which leverage spatial, sequential, or evolutionary relationships among amino acid residues. Tasks such as next-token prediction, fragment reordering, and predicting masked MSAs fall under this category.
    \item Cross-modal methods, which integrate information from multiple biological modalities—for instance, aligning sequence data with 3D structural information, evolutionary profiles, or even textual annotations—to learn joint representations that reflect multimodal biological knowledge.
\end{enumerate}

Overall, (self-)supervised pre-training has become a cornerstone in protein Transformer research, enabling improved generalization across a range of downstream tasks and opening new avenues for data-efficient modeling. As protein databases continue to grow in size and diversity, the role of SSL in scaling and advancing protein understanding will likely remain essential.

\subsection{Transformer Derivatives in Protein Informatics}\label{SecC}
Transformer models, originally developed for sequence-based tasks, have been adapted and extended to meet the specific needs of protein informatics. These derivatives maintain the core principles of the Transformer architecture, such as self-attention mechanisms and parallel processing, while incorporating modifications to enhance performance on tasks such as structure prediction, sequence-to-structure mapping, and functional annotation. This section provides an overview of the major Transformer derivatives developed for protein informatics, highlighting their key features, key references, and applications.
\subsubsection{Key Derivatives of the Transformer Architecture}

In order to provide an exhaustive overview, a comprehensive tabulation of the salient Transformer derivatives is presented in Appendix A (Table 1 and Table 2). This compilation delineates the pivotal attributes, pertinent application domains, and exemplary models of each Transformer variant, thereby offering a succinct synopsis conducive to an enhanced comprehension of their distinct roles and contributions within the realm of protein informatics.
\subsubsection{Future Directions for Transformer Derivatives}
The development of Transformer derivatives in protein informatics is an active area of research. Future work in this area is likely to focus on further optimizing these models for specific tasks, exploring new architectures that combine the strengths of different approaches, and addressing the challenges associated with computational efficiency and interpretability. Additionally, there is a growing interest in developing models that can handle multi-modal data, such as integrating protein sequences with structural information and other biological data types. These advancements are expected to lead to even more powerful and versatile models for protein informatics.
\subsection{Fundamentals of Protein Structure and Function}\label{SecD}
Proteins are fundamental macromolecules that play indispensable roles in virtually all biological processes, including catalysis, molecular transport, structural integrity, and signal transduction. The remarkable functional diversity of proteins arises from their distinct three-dimensional conformations, which are organized hierarchically into primary, secondary, tertiary, and quaternary structural levels. This section presents a systematic overview of these structural hierarchies, emphasizing the critical importance of protein folding and stability, the intricate relationship between structure and biological function, and the persistent challenges associated with accurately predicting protein structures.
\subsubsection{Hierarchical Organization of Protein Structure}

\begin{itemize}
    \item \noindent\textbf{Primary Structure:}
The primary structure refers to the linear sequence of amino acids linked by peptide bonds, which determines the folding trajectory and the ultimate three-dimensional conformation critical to protein function. Mutations at this level can disrupt folding and lead to dysfunction or disease \cite{a46}. Traditional methods such as sequence alignment and homology modeling have long been used to infer functional and evolutionary relationships \cite{a47} \cite{a48}. Transformer-based models have since enhanced primary structure analysis by capturing long-range dependencies, offering a deeper understanding of sequence-function relationships\cite{a49} \cite{a50}.

\item \noindent\textbf{Secondary Structure:}
Secondary structures, including $\alpha$-helices and $\beta$-sheets, arise from local hydrogen bonding patterns within the backbone. These elements provide mechanical stability and facilitate early folding events\cite{a51}. Experimental techniques such as X-ray crystallography and NMR have elucidated these motifs\cite{a52}, while classical predictors like Chou-Fasman and GOR rely on residue propensities\cite{a53} \cite{a54}. Modern Transformer architectures significantly improve prediction accuracy by modeling context-aware residue interactions\cite{a55}\cite{a56}.

\item \noindent\textbf{Tertiary Structure:}
Tertiary structure reflects the full 3D conformation of a polypeptide, integrating secondary elements and side-chain orientations to form functional domains, such as active and binding sites\cite{a57}. Prediction remains challenging due to the complexity of intramolecular forces. However, Transformer-based models—particularly AlphaFold—have revolutionized this task by learning spatial relationships from large-scale sequence–structure pairs\cite{a58} \cite{a59}.

\item \noindent\textbf{Quaternary Structure:} 
Quaternary structure entails the assembly of multiple polypeptide chains into functional protein complexes. This level is essential for the activity of many proteins, including hemoglobin and ribosomes. Accurate modeling of inter-subunit interactions is vital for drug discovery and structural annotation. Recent Transformer frameworks exhibit promising capabilities in predicting protein-protein interactions and complex formation\cite{a60} \cite{a61}.
\end{itemize}

\subsubsection{Protein Folding and Structural Stability}

Protein folding is an intrinsic and thermodynamically driven process through which a linear polypeptide chain acquires its specific three-dimensional native conformation, guided by intramolecular interactions among amino acid residues. This process adheres to the thermodynamic principle of free energy minimization, whereby the polypeptide traverses a conformational landscape—often described by the "folding funnel" model—to reach a low-energy functionally active state \cite{a62}. However, aberrations in the folding pathway can result in misfolding and subsequent aggregation, leading to pathological conditions such as Alzheimer's disease, in which misfolded proteins accumulate into non-functional aggregates that compromise cellular homeostasis \cite{a63}.

Protein stability denotes the capacity of a protein to preserve its native structure under varying physicochemical conditions, including fluctuations in pH, temperature, and ionic strength. Stability is modulated by intrinsic factors such as amino acid composition and structural interactions, as well as extrinsic environmental variables. While traditional approaches such as molecular dynamics simulations have been widely utilized to investigate protein stability, recent advancements have highlighted the growing efficacy of transformer-based models in predicting stability. These models leverage deep learning to extract and interpret sequence-structure relationships that underlie conformational robustness under stress conditions \cite{a64} \cite{a65}.

\subsubsection{The Relationship Between Protein Structure and Function}

\hspace*{1em}The structure of a protein is intricately tied to its biological function, as the specific spatial arrangement of amino acid residues determines the molecule’s ability to carry out precise biochemical tasks. Functional domains within proteins, such as catalytic or ligand-binding regions, are often evolutionarily conserved and serve as indicators of specific molecular activities. For instance, enzymatic proteins feature active sites with highly specialized geometries that facilitate substrate conversion, while immunoglobulins contain hypervariable regions designed to recognize and bind target antigens with high specificity.

Accurate prediction of protein function from sequence and structural data remains a central challenge in bioinformatics. Conventional methods typically employ homology-based inference, extrapolating function from proteins with known annotations and similar structures. In contrast, Transformer-based architectures—such as ProtTrans have demonstrated enhanced generalization across phylogenetically diverse sequences by leveraging large-scale pretraining on protein corpora. These models have shown promising capabilities in predicting functional sites, identifying subcellular localization patterns, and modeling biomolecular interactions with improved resolution \cite{a66} \cite{a67}.

\renewcommand{\arraystretch}{1.5}  % 行高是默认的1.5倍
\begin{table*}[ht]
\centering
\small
\caption{Overview of Selected Major Transformer Derivatives in Protein Informatics}
\begin{tabularx}{\linewidth}{>{\centering\arraybackslash}X X X >{\centering\arraybackslash}X}
\toprule
\textbf{Model Name} & \textbf{Key Features} & \textbf{Key Applications} & \textbf{Key References} \\
\midrule
AlphaFold & Hybrid architecture with Evoformer and Structure Module, and advanced attention mechanisms, and Multi-task training with structural supervision.& 3D structure prediction, and protein folding and sequence-to-structure mapping. & Jumper et al.\cite{a81} \\
ProtTrans & Pre-trained on large-scale protein sequences, and integration of sequence and evolutionary context. & Function prediction, and large-scale protein annotation and transfer learning for downstream tasks. & Elnaggar et al.\cite{a103}\\
RoseTTAFold & Computationally efficient, and end-to-end structure prediction. & Membrane protein structure prediction, and accelerated structural biology. & Baek et al.\cite{a82} \\
ESM-Fold & MSA-free architecture, and protein language model-based and High efficiency and scalability. & Genome-wide structural annotation, and orphan protein and rare species analysis. & Lin et al.\cite{a83}, Arana et al.\cite{a84}, Jumper et al.\cite{a85}, Senior et al.\cite{a86}\\
TAPE & Transferable protein embeddings, and transformer-based multi-task learning. & Protein-protein interaction (PPI) prediction, and low-resource fine-tuning scenarios. & Rao et al.\cite{a105} \\
DstruCCN & Hybrid architecture: CNN + Transformer, and binding site matrix-guided 3D reconstruction and superior performance on low-homology proteins. & Single-sequence protein structure prediction, and Low-homology or orphan protein modeling. & Zhou et al.\cite{a182} \\
ProteinBERT & BERT-based architecture adapted for proteins, and pretrained on large-scale protein datasets. & Enzyme classification and function annotation, and Ligand binding and subcellular localization prediction. & Brandes et al.\cite{a102} \\
Trans-MoRFs & High performance on IDR prediction, and self-attention for long-range dependency modeling. & MoRF identification in intrinsically disordered regions, functional annotation of dynamic protein regions.& Meng et al.\cite{a187} \\
ChemBERTa & BERT-based architecture adapted for chemical data, and fine-tuned on protein-ligand interaction datasets. & Early-stage virtual screening, and chemical property and interaction modeling. & Chithrananda et al.\cite{a107} \\
DeepChem & Integrates diverse deep learning models, including Transformers, and comprehensive drug discovery platform & Molecular property prediction for drug candidates, and flexible integration for customized workflows. & Ramsundar\cite{a181} \\
\bottomrule
\end{tabularx}
\end{table*}

\section{Transformers in Protein}\label{SecIII}
In recent years, Transformer models—originally developed for natural language processing—have been increasingly applied to protein informatics, leading to significant advancements across the field. By leveraging self-attention mechanisms, these models effectively capture complex dependencies within biological sequences, enabling breakthroughs in tasks such as prediction of protein structure and function prediction, protein–protein interaction modeling, and drug discovery. This section reviews the current progress and key applications of Transformer-based approaches in protein informatics, focusing on four major domains: protein structure prediction(\ref{SecA1}), protein function prediction(\ref{SecB1}), protein-protein interactions (PPIs)(\ref{SecC1}), and drug discovery(\ref{SecD1}).

\subsection{Transformers for Protein Structure Prediction}\label{SecA1}

Protein structure prediction remains one of the most critical challenges in computational biology, as it underpins our understanding of protein function, interactions, and potential for therapeutic intervention. For decades, researchers have relied on experimental methods such as X-ray crystallography and NMR spectroscopy, but these techniques are time consuming, costly, and require highly purified protein samples. As a result, computational methods for structure prediction have garnered significant attention, with deep learning methods, particularly Transformer models, playing an increasingly important role.

The application of Transformer models in protein structure prediction has significantly advanced the field by providing more accurate and efficient methods for predicting protein structures from sequences. These models have demonstrated remarkable performance in handling complex folds and long-range interactions, which are crucial for understanding protein function and interactions.

 1.\textbf{AlphaFold} \cite{a81}, developed by DeepMind, has revolutionized protein structure prediction by achieving near-experimental accuracy through an innovative deep learning framework. The model leverages a hybrid architecture combining an Evoformer (a Transformer-like module with axial attention) to analyze multi-sequence alignments and evolutionary couplings, and a Structure Module that refines atomic coordinates using geometric constraints. Key to its success are attention mechanisms, including triangular self-attention, which explicitly models 3D spatial relationships between residues while capturing long-range dependencies critical for folding. Trained on the Protein Data Bank (PDB) with a multi-task loss function incorporating distograms and physical constraints, AlphaFold attained unprecedented performance in CASP14 (2020), achieving a median Global Distance Test (GDT) score of 92.4. Beyond static structures, subsequent releases like AlphaFold-Multimer extended predictions to protein complexes, and the AlphaFold DB provided open access to over 200 million structures, transforming fields from drug discovery to synthetic biology. Despite limitations in modeling dynamic conformations or disordered regions, AlphaFold’s open-source release has democratized structural biology, inspiring tools like RoseTTAFold and ESMFold, and underscoring the potential of AI to accelerate scientific discovery.
 
 2.\textbf{RoseTTAFold} \cite{a82}, another notable model developed by the University of Washington’s Rosetta group, uses a similar approach to AlphaFold but is designed to be more computationally efficient . RoseTTAFold builds upon the Transformer architecture by combining it with a graph-based approach, which enables it to predict protein structures faster without sacrificing accuracy. This model has been particularly valuable for solving structures of membrane proteins, which have historically been difficult to predict with traditional methods. Both AlphaFold and RoseTTAFold highlight the significant potential of Transformer models to accelerate the process of structural biology, facilitating drug discovery and therapeutic development. 

3.\textbf{ESM-Fold}, developed by Meta AI, represents a significant innovation in protein structure prediction by employing language models trained on extensive protein sequence datasets\cite{a83}. Unlike traditional methods that rely heavily on multiple sequence alignments (MSA) to infer evolutionary relationships and structural features, ESM-Fold bypasses the need for MSA entirely, extracting meaningful information directly from single sequences. This approach significantly reduces computational requirements and avoids the bottlenecks associated with generating alignments\cite{a83}. Trained on billions of protein sequences, ESM-Fold captures the intricate contextual relationships between amino acids, enabling it to predict the three-dimensional structure of proteins with remarkable accuracy, even for sequences with minimal or no homology to known structures\cite{a83} \cite{a84}. Its efficiency makes it highly suitable for high-throughput studies, the analysis of newly sequenced genomes, and applications in synthetic biology\cite{a85}. ESM-Fold also excels at predicting structures for orphan proteins or those from understudied organisms, where homologous sequence data may be scarce.By leveraging the power of language models, ESM-Fold reduces computational demands and broadens the scope of proteins that can be effectively analyzed, accelerating advancements in protein science and its applications in drug discovery and bioengineering\cite{a83} \cite{a86}.

4.\textbf{OmegaFold}, developed to meet the growing demand for efficient and accurate protein structure prediction, represents a breakthrough in utilizing Transformer-based architectures for real-time analysis\cite{a180}. Unlike traditional models that depend on multiple sequence alignments (MSA) to derive evolutionary relationships, OmegaFold operates directly on single protein sequences. This streamlined approach significantly reduces computational requirements and enhances applicability, particularly in high-throughput studies and scenarios where homologous sequence data is limited. OmegaFold incorporates advanced Transformer-based mechanisms to capture both local and long-range dependencies within protein sequences, enabling it to predict three-dimensional structures with high accuracy. By bypassing the MSA step, OmegaFold avoids the computational bottlenecks associated with alignment generation, making it ideal for rapid analyses of newly sequenced proteins or proteins from diverse and uncharacterized organisms. OmegaFold's exceptional performance on orphan proteins and novel sequences, and its ability to generalize across various protein families, make it a versatile tool for high-throughput studies, including drug discovery and synthetic biology applications. In validation studies, OmegaFold has shown competitive accuracy compared to leading models, including AlphaFold and RoseTTAFold, particularly for single-sequence inputs. Overall, OmegaFold offers a transformative approach to protein structure prediction by leveraging the power of Transformer-based models while eliminating the reliance on MSA. Its combination of speed, accuracy, and adaptability makes it an essential resource for researchers tackling the challenges of modern structural biology\cite{a180}.

5.\textbf{ProtGPT2}, inspired by the success of generative pretraining in natural language processing (e.g., GPT models), applies similar methodologies to protein informatics\cite{a179}. Designed to generate novel protein sequences that are both structurally viable and functionally diverse, ProtGPT2 leverages a Transformer-based architecture optimized for sequence generation. Trained on extensive protein sequence databases, the model captures the complex statistical relationships between amino acids, enabling it to propose de novo sequences with realistic biophysical properties, as confirmed through computational validations such as secondary structure prediction and solvent accessibility analysis. A key advantage of ProtGPT2 is its focus on structural and functional diversity, generating a wide array of sequences that enhance the exploration of sequence-function landscapes. This diversity is particularly valuable in applications like directed evolution, where a diverse library can increase the chances of identifying sequences with desired characteristics. ProtGPT2 also integrates structural constraints into its generative process, ensuring that the proposed sequences are compatible with known folding patterns, minimizing the risk of aggregation or misfolding. Its ability to design sequences for entirely novel protein folds makes it a powerful tool for tackling challenges that cannot be addressed by template-based methods. Applications of ProtGPT2 extend to biocatalyst development and therapeutic protein design, where it accelerates the engineering cycle by reducing the need for extensive experimental screening\cite{a89}. ProtGPT2 exemplifies the transformative potential of generative pretraining in protein sequence design, positioning it as a cornerstone technology for future innovations in protein engineering and synthetic biology.

6.\textbf{RFDiffusion}, developed by the Baker Lab at the University of Washington, applies diffusion modeling—originally used in image generation—to de novo protein design by generating novel backbones through a forward-backward noise process\cite{a100}. Unlike traditional sequence- or template-based methods, it reconstructs viable structures from noise, enabling exploration of conformational space under physical and biochemical constraints. The model excels in producing experimentally viable proteins, supported by geometric, interaction, and evolutionary scoring. Applications span therapeutic design, enzyme engineering, and scaffold generation, though challenges remain in computational cost and constraint accuracy. As a pioneering integration of generative modeling and structural biology, RFDiffusion marks a significant advance in computational protein engineering.

7.\textbf{GraphTrans}, a model integrating graph neural networks (GNNs) with Transformer layers, enhances protein complex prediction and structure refinement by capturing both spatial residue relationships and long-range sequence dependencies \cite{a101}. GNNs model topological interactions within protein complexes, while Transformers enrich global context understanding, enabling GraphTrans to predict and iteratively refine complex structures with high accuracy. This synergy allows the model to correct spatial inaccuracies and better represent multi-subunit interactions, highlighting its value in advancing protein structure analysis and functional annotation.

8. In recent years, the integration of diverse deep learning architectures has emerged as a promising strategy to enhance the accuracy and robustness of protein structure prediction. Zhou et al. (2024) proposed a novel model named \textbf{DstruCCN}, which combines Convolutional Neural Networks (CNNs) and a supervised Transformer-based protein language model to perform single-sequence protein structure prediction\cite{a182}. The CNN component focuses on capturing local spatial dependencies among residues, while the Transformer module models long-range sequence dependencies by leveraging its self-attention mechanism. The features extracted by both networks are fused to predict the binding site matrix of proteins, which is then used to reconstruct the three-dimensional (3D) structure through energy minimization-based refinement.

The proposed approach outperforms traditional CNN-only or Transformer-only architectures, particularly in modeling complex protein folds with sparse evolutionary information. This is especially relevant for orphan proteins or sequences with limited homologs, where traditional MSA-based methods may underperform. Moreover, the model's modularity allows for flexible integration into broader protein structure prediction pipelines. The success of DstruCCN highlights the potential of hybrid deep neural networks in advancing single-sequence protein modeling, which is an increasingly important direction in post-AlphaFold research.

9. In the evolving landscape of protein secondary structure prediction (PSSP), the integration of diverse deep learning architectures has shown significant promise. Chen et al. (2024) introduced \textbf{MFTrans}, a novel deep learning-based multi-feature fusion network designed to enhance the precision and efficiency of PSSP\cite{a183}. This model employs a Multiple Sequence Alignment (MSA) Transformer in combination with a multi-view deep learning architecture to effectively capture both global and local features of protein sequences. MFTrans integrates diverse features generated by protein sequences, including MSA, sequence information, evolutionary information, and hidden state information, using a multi-feature fusion strategy. The MSA Transformer is utilized to interleave row and column attention across the input MSA, while a Transformer encoder and decoder are introduced to enhance the extracted high-level features. A hybrid network architecture, combining a Convolutional Neural Network (CNN) with a bidirectional Gated Recurrent Unit (BiGRU) network, is used to further extract high-level features after feature fusion.

In independent tests, experimental results demonstrate that MFTrans has superior generalization ability, outperforming other state-of-the-art PSSP models by 3\% on average on public benchmarks including CASP12, CASP13, CASP14, TEST2016, TEST2018, and CB513 . Case studies further highlight its advanced performance in predicting mutation sites. The success of MFTrans underscores the potential of integrating multiple feature representations and hybrid neural network architectures in advancing the field of protein secondary structure prediction.

10. Recent advancements in deep learning have spurred the development of hybrid architectures that effectively integrate complementary modeling strategies for improved protein structure prediction. One such approach is \textbf{TransConv}(2025), a model that fuses transformer-based attention mechanisms with convolutional neural networks to enhance the accuracy of secondary structure prediction\cite{a185}. While transformers excel at capturing long-range dependencies across protein sequences, convolutional layers are adept at extracting local features critical to identifying structural motifs. By combining these two paradigms, TransConv achieves a balanced representation of both global and local sequence-context relationships. This integrated framework enables the model to better capture structural patterns derived from backbone hydrogen bonding interactions, a key determinant of secondary structure. Experimental evaluations on standard benchmark datasets have demonstrated that TransConv consistently outperforms several state-of-the-art methods, highlighting its potential as a robust tool for efficient and accurate secondary structure prediction.

11. Beyond conventional sequence-based modeling, recent research has demonstrated the efficacy of transformers in reconstructing atomic protein structures from cryo-electron microscopy (cryo-EM) density maps. One representative work introduces \textbf{Cryo2Struct} (2024), a fully automated framework that employs a 3D transformer network to identify atomic coordinates and amino acid types directly from volumetric cryo-EM data\cite{a186}. To convert these predictions into coherent backbone conformations, the model incorporates a Hidden Markov Model (HMM) that sequentially connects the detected atoms, thereby forming complete protein backbones. This end-to-end pipeline enables template-free de novo structure prediction, a crucial advancement for modeling proteins lacking homologous or predicted templates. Compared to traditional ab initio methods such as Phenix, Cryo2Struct demonstrates superior accuracy and completeness, even across varying density resolutions and protein sizes. This approach highlights the versatility of transformer models in bridging experimental data with computational inference, pushing forward the frontiers of structural bioinformatics.

\begin{figure*}[htbp]
    \centering
    \includegraphics[width=\textwidth]{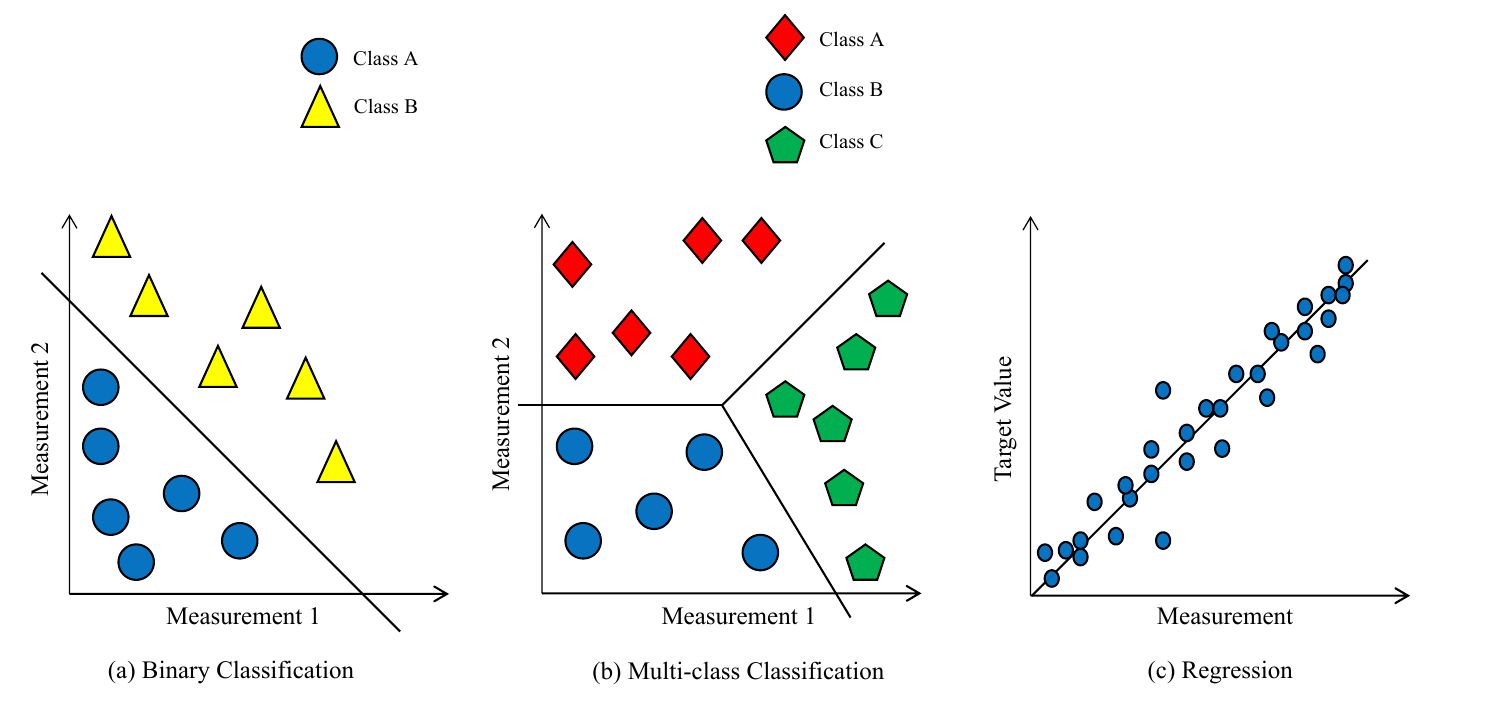}
    \vspace{-10pt} % 
    \caption{Illustrative diagram of three fundamental supervised learning tasks.
Supervised learning in machine learning (ML) is typically categorized into classification and regression tasks. While two-dimensional representations are used here for conceptual clarity, real-world datasets often reside in high-dimensional feature spaces.
(a) In binary classification, each sample belongs to one of two possible categories. For instance, a model may classify protein variants as either stable or unstable\cite{a199}, or determine whether a protein is a G-protein-coupled receptor or not, based on sequence-derived features and machine learning models\cite{a200}.
(b) Multi-class classification involves assigning samples to one of several discrete classes. For example, recent studies have developed machine learning models to predict the subcellular localization of human proteins—such as nucleus, cytoplasm, mitochondria, and extracellular regions—based on features extracted from immunohistochemistry images\cite{a201}.
(c) In regression tasks, the goal is to predict continuous numerical properties of proteins. For instance, recent models have been developed to estimate protein solubility levels directly from sequence-derived or structural features, enabling fine-grained prediction beyond binary soluble/insoluble classification\cite{a202}.}
    \label{fig:enter-label2}
\end{figure*}

\subsection{Transformers for Protein Function Prediction}\label{SecB1}

Prediction of protein function is crucial to understand the role of proteins in cellular processes and their potential as drug targets. Traditional methods for prediction of protein functions are based heavily on sequence homology and annotations from protein databases. However, these methods are unable to predict functions for proteins that lack close homologs or belong to uncharacterized families.

\begin{figure}[ht]
    \centering
    \includegraphics[width=1.0\linewidth]{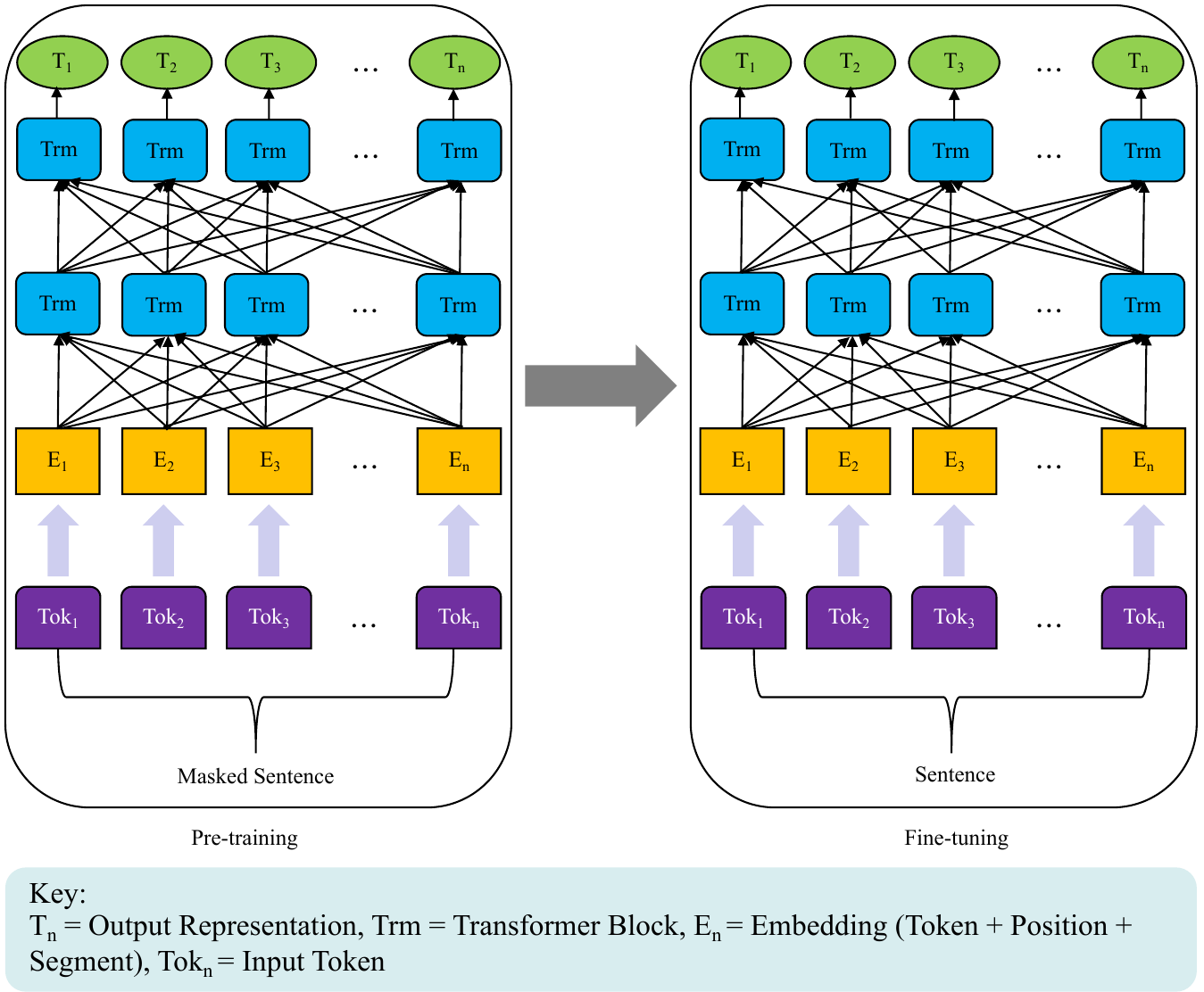} % 75%当前栏宽度
    \caption{
        The structure of the BERT model. 
        This figure shows the Bidirectional Encoder Representations from Transformers (BERT) architecture, 
        highlighting both pre-training and fine-tuning phases.
    }
    \label{fig:bert_structure}
\end{figure}

\textbf{1.ProteinBERT.} Recent advancements in Transformer-based models have significantly enhanced protein function prediction. One notable example is ProtBERT \cite{a102}, a model adapted from the BERT architecture. 
Fig.~\ref{fig:bert_structure} illustrates the architecture of the original BERT model, which ProteinBERT inherits and adapts for protein sequences.
ProteinBERT is pre-trained on large-scale protein sequence datasets and learns to capture contextual dependencies between amino acids. When fine-tuned for specific tasks such as enzyme classification, ligand binding prediction, or subcellular localization, ProteinBERT achieves state-of-the-art performance.

A key strength of ProteinBERT lies in its ability to encode rich semantic representations of protein sequences. By learning position-aware, context-sensitive embeddings, it can identify conserved motifs and functionally relevant regions even in remote homologs. This enables it to predict functions for novel or poorly annotated proteins that traditional alignment-based tools often miss.

\textbf{2.ProtTrans.} Building on this foundation, ProtTrans \cite{a103} pushes the boundary further by leveraging larger model architectures and broader training corpora, including billions of protein sequences. ProtTrans is capable of predicting diverse protein properties such as stability, binding affinity, and molecular function. By integrating sequence-based information with evolutionary context, it provides robust predictions even for proteins with limited annotation. This makes ProtTrans a powerful tool for large-scale protein annotation in genomic studies.

Unlike earlier models, ProtTrans comprises a collection of Transformer architectures, including BERT, XLNet, Albert, and T5—each trained on massive datasets such as UniRef100 and BFD, encompassing billions of protein sequences. These models are trained using self-supervised learning objectives, particularly masked language modeling and sequence-to-sequence prediction. This enables them to learn contextual and evolutionary patterns in protein sequences without requiring labeled data.

The key innovation in ProtTrans lies in its ability to generalize across diverse protein families by capturing both local motifs and global structural signals embedded in the sequence. The use of multiple model types further enhances its flexibility across tasks, as each architecture offers unique advantages in handling long-range dependencies, directional context, or compression efficiency. Additionally, ProtTrans exhibits strong transfer learning capabilities, performing well on downstream tasks even with limited training examples, which is particularly valuable in domains with sparse functional annotations.

3. In recent years, Transformer-based models have also shown considerable potential in the domain of protein function analysis, particularly in identifying biologically important regions such as binding sites.One notable example is \textbf{Deep-ProBind}(2025), a hybrid model designed to predict protein-binding peptides by integrating both sequence-derived and structural features\cite{a184}.This approach utilizes Bidirectional Encoder Representations from Transformers (BERT) alongside a Pseudo Position-Specific Scoring Matrix transformed via Discrete Wavelet Transform (PsePSSM-DWT), enabling the extraction of complex sequence patterns and evolutionary characteristics.A key contribution of Deep-ProBind is the incorporation of the SHapley Additive exPlanations (SHAP) algorithm, which selects informative hybrid features prior to classification using a deep neural network.In extensive benchmark testing, the model achieved an accuracy of 92.67\% using tenfold cross-validation, and 93.62\% on independent datasets, outperforming traditional machine learning methods and other existing models by a notable margin.Although Deep-ProBind is not a structural prediction tool per se, its strategic use of structural descriptors underscores the broader utility of transformer-based models in understanding protein behavior and interactions, which is critical for pharmacological applications and peptide drug discovery.

4. In recent years, growing attention has been directed toward understanding the functional roles of intrinsically disordered regions (IDRs) in proteins, especially their short interaction-prone segments known as molecular recognition features (MoRFs). Accurately identifying MoRFs remains a computational challenge due to their disorder-to-order transitions and the limited availability of experimentally validated annotations. To overcome these limitations, a recent study proposed \textbf{Trans-MoRFs} (2025), a novel predictor based on the transformer architecture\cite{a187}. Trans-MoRFs leverage self-attention mechanisms to effectively capture long-range dependencies within protein sequences, making them particularly robust across proteins of varying lengths. The model demonstrated strong predictive performance, achieving a mean area under the curve (AUC) of 0.94 on multiple benchmarks, and significantly outperformed existing MoRF predictors across several evaluation metrics. This work exemplifies the expanding role of transformer models in structural and functional protein annotation, extending their applicability beyond structured domains to disordered and dynamic regions, thereby enhancing our understanding of protein function and supporting drug target discovery.

5. Recent advances in protein function prediction have begun to emphasize not only predictive performance but also interpretability, particularly when using complex models such as transformers. A notable example is the study by Wenzel et al.\cite{a188}(2024), which investigates the internal mechanisms of transformer models fine-tuned for protein function tasks, including Gene Ontology (GO) term and Enzyme Commission (EC) number prediction. The authors extended the widely used explainable AI (XAI) method, Integrated Gradients, to analyze not only input attributions but also latent representations within transformer architectures. Their approach allowed for the identification of amino acid residues that the model attends to most strongly, revealing biologically meaningful patterns. Specifically, they demonstrated that certain transformer heads consistently correspond to known functional regions, such as transmembrane domains and enzymatic active sites. This work highlights the growing potential of interpretable transformer models in uncovering biologically relevant insights from sequence data, thereby enhancing both the trustworthiness and utility of deep learning in protein function annotation.

6. Traditional graph-based methods may struggle to represent these multi-hop relationships, limiting the extraction of deeper functional signals. To address this, recent advances have explored the integration of graph serialization with Transformer-based architectures.

\textbf{SEGT-GO}(2025)\cite{a197} introduces a novel framework that transforms multi-hop neighborhoods within PPI networks into serialized representations, allowing a Graph Transformer to learn latent functional dependencies across the interaction space. By converting local and global network topologies into sequences of embeddings, the model capitalizes on the attention mechanism's ability to model non-local relationships. In addition, the framework incorporates SHAP-based explainable AI techniques, leveraging game-theoretic principles to refine input relevance and reduce noise, thereby enhancing interpretability and robustness.

Experimental evaluations across both large-scale and limited-size datasets demonstrate SEGT-GO's superiority over previous state-of-the-art models such as DeepGraphGO. Notably, the method shows strong generalization capabilities in cross-species scenarios and when predicting functions of previously unseen proteins, highlighting the potential of combining Transformer-based graph modeling with explainable AI in scalable protein function analysis.

7. In recent years, Transformer-based approaches have increasingly demonstrated their utility in automating complex tasks within protein function prediction, a domain traditionally hindered by the high cost and low scalability of experimental methods. One recent contribution is \textbf{MAGO} (2025)\cite{a198}, a model that combines the representational power of Transformer architectures with the optimization capabilities of Automated Machine Learning (AutoML) to improve large-scale protein annotation. Unlike earlier machine learning pipelines, MAGO autonomously selects and tunes model configurations while leveraging contextual embeddings extracted from raw protein sequences. Furthermore, the enhanced variant MAGO+ incorporates evolutionary signals via integration with BLASTp, effectively unifying deep learning-based feature learning and alignment-based similarity. Benchmark evaluations reveal that both MAGO and MAGO+ surpass several state-of-the-art approaches in Fmax and other performance metrics, with improvements found to be statistically significant. These results not only reaffirm the value of attention mechanisms for capturing functional patterns in biological sequences but also illustrate how AutoML can systematically refine deep learning frameworks for robust cross-protein generalization, paving the way for more adaptive and interpretable solutions in computational proteomics.
\subsection{Transformers for Protein-Protein Interactions (PPIs)}\label{SecC1}

Protein-protein interactions (PPIs) are fundamental to nearly all cellular activities, including signal transduction, gene regulation, and immune response. Accurately predicting PPIs is essential for drug discovery, understanding disease mechanisms, and identifying therapeutic targets. While experimental methods such as yeast two-hybrid screening and co-immunoprecipitation are widely used, they are resource-intensive, time-consuming, and limited in scalability.

To overcome these limitations, deep learning, especially Transformer-based models, has been increasingly applied to computational PPI prediction. These models leverage sequence and structural information to identify interaction patterns and functional relationships between proteins.

\textbf{1.DeepPPI: Sequence-based attention for interaction prediction}
DeepPPI\cite{a104} is a deep neural network model that predicts protein interactions based solely on amino acid sequences. It integrates an attention mechanism that enables the model to focus on biologically relevant sequence motifs and potential interaction regions. This targeted attention improves prediction accuracy, particularly in cases where structural information is unavailable. One of the key advantages of DeepPPI is its ability to generalize across proteins with low sequence similarity, making it robust for large-scale proteomic analyses. However, its reliance on primary sequence data alone may limit performance when structural features play a dominant role in mediating interactions.

\textbf{2.TAPE: Multi-task Transformer framework}
The TAPE (Task-Aware Protein Embedding) framework\cite{a105} builds on a Transformer-based architecture with a multi-task learning setup. It is pretrained on over 30 million protein sequences, learning generalized embeddings that can be fine-tuned for specific tasks, including PPI prediction. TAPE’s strength lies in its ability to transfer knowledge across tasks, significantly reducing the need for large labeled datasets in downstream applications. The model captures complex sequence patterns and functional relationships, which enhances performance in diverse biological prediction tasks. Nonetheless, without explicit incorporation of structural data, TAPE may struggle to model spatial constraints that are critical for certain PPIs.

\textbf{3.Graph Attention Networks (GATs): Structure-aware interaction modeling}
Graph-based approaches like Graph Attention Networks (GATs)\cite{a106} treat proteins as graphs, where amino acids are nodes and edges represent spatial or functional relationships. The attention mechanism in GATs assigns different importance weights to neighboring nodes, enabling the model to emphasize critical regions such as binding pockets or conserved structural motifs. This structure-aware modeling makes GATs particularly effective in capturing 3D context and non-linear interaction pathways, which are essential for accurately predicting interactions in complex molecular environments. However, GATs often require high-quality structural input, and the computational cost associated with large protein graphs can be substantial.

4. Recent efforts in protein–protein interaction site (PPIS) prediction have explored hybrid models that combine structural and sequential features for enhanced accuracy. A notable contribution in this direction is the \textbf{GACT-PPIS}(2024) model, which integrates an enhanced graph attention mechanism with deep transformer networks to predict interaction sites on protein surfaces\cite{a189}. Specifically, the model utilizes an Enhanced Graph Attention Network (EGAT) enriched with residual and identity mappings, in conjunction with a transformer encoder to capture long-range dependencies in amino acid sequences. Graph Convolutional Networks (GCNs) are also employed to aggregate local neighborhood information from the protein’s structural graph. This multi-level fusion of spatial and sequential data enables GACT-PPIS to outperform several state-of-the-art models across multiple benchmark datasets, including Test-60 and UBTest-31-6, showing superior performance in terms of Recall, F1-score, AUROC, AUPRC, and overall generalization. The success of GACT-PPIS highlights the value of combining graph-based and transformer-based representations for precise residue-level interaction prediction.

5. Recent advancements have also explored the application of Transformer-based models in predicting protein–protein interaction (PPI) binding sites, a critical aspect of understanding cellular mechanisms and therapeutic target identification. A representative study introduced \textbf{TranP-B-site} (2025), a model that utilizes Transformer-generated embeddings to encode amino acid sequence information, combined with convolutional neural networks (CNNs) to detect PPI binding sites\cite{a190}. The model extracts two types of features for each residue—one-hot encodings for local patterns and Transformer-based embeddings for global context—and processes them using a hybrid architecture. Specifically, a windowing strategy is employed to capture localized information, which is then separately processed by CNN and fully connected layers before final classification. This dual-pathway design enables the model to integrate both fine-grained and holistic representations effectively. TranP-B-site demonstrated notable improvements over previous state-of-the-art sequence-based PPI models, achieving significant gains in accuracy and Matthews Correlation Coefficient (MCC) on independent test datasets. Furthermore, the model exhibited robust performance on a newly curated dataset derived from the PDB, indicating its generalization capability across diverse protein sequences. This work underscores the utility of Transformer embeddings in enhancing predictive power for interaction site identification and highlights a promising direction for future PPI modeling.

6. Uncertainty estimation has recently become a critical factor in enhancing the reliability of protein–protein interaction (PPI) predictions using Transformer-based models. One recent contribution in this area is the \textbf{TUnA} framework(2024), which integrates uncertainty modeling into sequence-based PPI prediction\cite{a191}. TUnA leverages Transformer encoders in combination with ESM-2 protein embeddings to extract informative representations of amino acid sequences. Notably, it incorporates a Spectral-normalized Neural Gaussian Process to quantify prediction uncertainty, enabling the model not only to make accurate predictions but also to estimate the confidence level of its outputs. This is particularly valuable for applications involving previously unseen proteins, where conventional models often struggle to generalize. TUnA has demonstrated state-of-the-art performance on benchmark datasets and showed that its uncertainty-aware predictions could effectively filter out unreliable outputs, thereby reducing false positives and narrowing the gap between in silico predictions and experimental validation. This approach illustrates a promising direction in building interpretable and trustworthy Transformer-based frameworks for PPI prediction.

7. Recent research has also explored the use of Transformer architectures for predicting protein-protein binding affinity (BA), a key aspect of understanding molecular interactions. One study conducted a comparative analysis of convolutional neural networks (CNNs) and Transformer-based models to evaluate their effectiveness in predicting binding affinity using only protein sequence data(2024)\cite{a192}. The models were designed to output Gibbs free energy changes, which serve as a quantitative measure of interaction strength between protein pairs.

In this work, multiple model variants were developed using both TensorFlow and PyTorch, including a Transformer-based model built upon ProteinBERT. The CNN architecture was effective at capturing local sequence features, while the Transformer model leveraged self-attention mechanisms to learn long-range dependencies within protein sequences. Different sequence encoding techniques were used, including one-hot encoding, sequence-statistics-content, and position-specific scoring matrices.

The results demonstrated that both architectures could achieve comparable predictive accuracy, although they presented distinct trade-offs. The CNN model could process full-length sequences but required significantly more preprocessing time. In contrast, the Transformer model maintained competitive accuracy with less computational overhead, though it excluded very long sequences due to input length limitations. This study underscores the growing potential of Transformer-based models in capturing complex sequence-level features for protein-protein interaction prediction and binding affinity estimation.

\subsection{Transformers for Drug Discovery and Target Identification}\label{SecD1}

Transformer-based models have recently revolutionized drug discovery by enabling more accurate virtual screening, predicting protein-ligand binding affinities, and supporting de novo drug design. Their ability to learn from large-scale chemical and biological datasets allows for efficient prediction of interactions between small molecules and protein targets, significantly reducing the cost and time associated with traditional experimental methods.

In recent years, Transformer-based models have rapidly gained prominence in the domain of drug discovery, particularly for their ability to handle complex and heterogeneous data. A comprehensive review by Jian Jiang et al. (2024) \cite{a193}emphasizes the transformative role of Transformers in various aspects of the drug discovery pipeline, including drug target identification, molecular design, and property prediction. The authors highlight that the hierarchical attention mechanisms inherent to Transformer architectures enable them to effectively model sequential dependencies in biological and chemical data. These models have shown strong adaptability when pre-trained on large-scale datasets, allowing transferability across tasks such as virtual screening, lead optimization, and even protein engineering. Notably, their interdisciplinary application across biology, chemistry, and pharmacology facilitates an integrative approach to discovery, bridging the gaps between domains. The review also outlines emerging directions, including the use of Transformers for single-cell data analysis, chemical language modeling, and biological image interpretation, underscoring their growing influence in drug discovery and biomedical research more broadly.

\textbf{1.ChemBERTa} is a prominent model in this area, fine-tuned on protein-ligand interaction datasets. Leveraging a BERT-style architecture, it captures complex patterns in chemical structures and their binding behaviors. ChemBERTa has demonstrated superior performance compared to traditional methods like molecular docking and molecular dynamics simulations in predicting binding affinities. This enables faster and more cost-effective early-stage drug screening by reducing reliance on time-consuming lab experiments \cite{a107}.

\textbf{2.MolBERT} utilizes Transformer models to learn molecular representations from SMILES (Simplified Molecular Input Line Entry System) strings. Pretrained on large-scale molecular datasets, it can predict not only binding affinity with protein targets but also toxicity and efficacy profiles of candidate compounds. This enables MolBERT to serve as a versatile tool throughout the drug development pipeline, supporting both safety assessment and lead optimization \cite{a108}.

\textbf{3.De novo Drug Design with Graph Transformers:} In addition to these models, Transformer-based approaches have been used for de novo drug design, an innovative method for generating entirely new molecules with specific binding properties. Graph-based models combined with Transformer architectures enable the generation of novel molecular structures by predicting optimal atom-to-atom connections for desired binding properties. This approach not only speeds up the drug design process but also allows for a more efficient exploration of chemical space compared to traditional drug design methods \cite{a109}.

\textbf{4.MolFormer} represents a next-generation Transformer model aimed at understanding the three-dimensional (3D) conformations of drug-like molecules. It predicts how small molecules adopt specific geometries within protein binding sites, aiding in the rational design of compounds with favorable spatial orientation and interaction properties. MolFormer is particularly valuable in refining molecular structures during the early stages of lead optimization \cite{a110}.

\textbf{5.DeepChem} is a machine learning platform that integrates various deep models, including Transformer-based architectures, for comprehensive drug discovery tasks. It supports the prediction of multiple molecular properties such as solubility, toxicity, and binding affinity. By combining rich chemical representations with advanced learning techniques, DeepChem enables rapid prioritization of promising drug candidates\cite{a181}.

\textbf{6.RL-based Transformer Models for Molecule Generation:} Reinforcement Learning-based models combined with Transformers have been applied to drug design, where a model learns the optimal drug molecule design through iterative feedback. These models leverage reinforcement learning strategies to iteratively generate new molecules that not only bind effectively to the target protein but also optimize other properties such as solubility and toxicity. By using a reward-based system, these models can effectively explore the vast chemical space and suggest novel molecules for drug development.

7. Recent advancements in drug discovery have utilized transformer-based models for enhanced molecular property prediction. A notable approach is the BERT-based model for HIV replication inhibition, which fine-tunes the transformer architecture on the MoleculeNet-HIV dataset represented by SMILES notation. By leveraging BERT's ability to capture intricate molecular patterns, the model achieves high prediction accuracy and strong generalization capabilities\cite{a194}(2024). This method demonstrates significant potential for accelerating drug discovery, reducing both time and cost, and offering a promising solution for more efficient target identification and molecular design.

8. Recent advances in Transformer-based generative models have opened new avenues for de novo drug design, especially when integrated with optimization strategies targeting multiple pharmaceutical objectives. One notable approach proposes a comprehensive framework that combines latent Transformer architectures with a many-objective optimization pipeline\cite{a195}(2024). This system jointly considers key pharmacokinetic and pharmacodynamic criteria—such as absorption, distribution, metabolism, excretion, toxicity (ADMET), and molecular docking—within a multi-objective design strategy.

In particular, two latent Transformer-based models, ReLSO and FragNet, were evaluated for their ability to encode and reconstruct molecular structures. ReLSO demonstrated superior performance, offering more coherent latent space organization and reconstruction accuracy. Building upon this, the framework incorporated six different many-objective metaheuristic algorithms—including those based on evolutionary principles and particle swarm optimization—to explore candidate compounds targeting human lysophosphatidic acid receptor 1 (LPA1), a protein implicated in cancer.

Among the tested strategies, a decomposition-based evolutionary algorithm yielded the most favorable balance across multiple drug design objectives, achieving high binding affinity, low toxicity, and strong drug-likeness. This work exemplifies the synergy between Transformer-driven molecular generation and computational intelligence methods for multi-criteria optimization, highlighting a scalable path toward more robust and realistic drug candidate discovery.

9. The emergence of the Transformer architecture has profoundly influenced computational chemistry, enabling a paradigm shift in how molecular data is interpreted, modeled, and utilized in drug discovery. Inspired by parallels between chemical notation and natural language, researchers have successfully applied language modeling techniques to challenges such as retrosynthetic analysis, molecular generation, and exploration of vast chemical spaces. Initial approaches focused on task-specific models trained on linear molecular representations (e.g., SMILES), but recent developments have expanded these capabilities to incorporate diverse modalities—including spectroscopic data, synthetic pathways, and even human-readable language inputs.

This evolution has culminated in the application of large language models (LLMs), which offer a unified framework for solving a broad range of chemistry-related problems. By leveraging the flexibility and expressiveness of natural language, LLMs are increasingly being used to integrate multi-source data and generalize across chemical tasks. These advancements point toward a future where machine learning models serve as adaptive and intelligent agents in the drug discovery pipeline, capable of synthesizing knowledge across domains and accelerating decision-making at every stage \cite{a196}(2024).

These advancements demonstrate the power of Transformer-based models in revolutionizing drug discovery by improving the prediction of protein-ligand interactions, enabling de novo design of novel drug candidates, and refining the properties of existing molecules. With their ability to process large-scale datasets and predict complex molecular behaviors, these models are accelerating the development of new therapeutics, bringing us closer to more effective treatments in less time.

\section{Advantages \& Challenges of Transformer Models}

\subsection{Advantages of Transformers in Protein Informatics}

Transformer models have profoundly reshaped the landscape of protein informatics, offering a set of compelling advantages over traditional computational approaches.

\textbf{1.Capturing Long-Range Dependencies:}

The self-attention mechanism of Transformers enables them to model long-range interactions in protein sequences, which are crucial for tasks such as tertiary and quaternary structure prediction. Unlike recurrent neural networks (RNNs), Transformers analyze all sequence elements simultaneously, efficiently capturing context from distant residues\cite{a38}.

\textbf{2.Scalability and Pretraining Benefits:}

Transformer models, especially in their pre-trained forms (e.g., ProtTrans, ESM), are scalable to massive datasets, providing a foundation for transfer learning across various protein tasks. This is advantageous when computational resources are limited or when training data for specific protein tasks are scarce \cite{a39} \cite{a40}.

\textbf{3.Parallel Computation and Efficiency:}

Transformers allow for parallelized computation during training and inference due to their non-recurrent architecture, significantly reducing training time. This scalability enables their use in high-throughput tasks like structural screening and synthetic protein design.

\textbf{4.Multimodal Integration:}

Advanced Transformer derivatives like AlphaFold2 and ESM-3 can integrate multiple types of biological information, including evolutionary profiles, structural constraints, and biochemical features, enabling more holistic and accurate predictions.

\textbf{5.Generalizability Across Diverse Protein Families:}

Transformers trained on large, diverse protein datasets are capable of generalizing to unseen sequences, including those from under-characterized or evolutionarily distant organisms. This makes them especially valuable for functional annotation in metagenomics and novel protein discovery\cite{a41}.

\subsection{Challenges of Applying Transformers in Protein Research}

Despite their advantages, Transformer models face several persistent challenges that limit their broader adoption and utility.

\textbf{1.Computational Resource Intensity:}

Scalability Issues: Transformer models are inherently resource intensive due to their quadratic complexity with respect to the sequence length in the self-attention mechanism. For protein sequences, which can be thousands of amino acids long, this computational demand becomes a significant bottleneck \cite{a110}. Training models like AlphaFold2 require high-performance computing clusters with specialized hardware such as GPUs or TPUs, making it inaccessible to many researchers and institutions with limited resources.

Memory Consumption: The large number of parameters in these models leads to substantial memory consumption. For example, models such as ProtT5-XL-UniRef50 have more than 3 billion parameters, which requires advanced memory management techniques and hardware with large memory capacities \cite{a112}.

Inference Time: The time required for inference can also be prohibitive, especially when processing large datasets or performing real-time analysis. This limitation affects the practical deployment of Transformer models in applications such as high-throughput drug screening, where speed is crucial \cite{a114}.

Energy Consumption and Environmental Impact: The extensive computational resources required not only increase costs but also have environmental implications due to high energy consumption, raising concerns about the sustainability of using such models at scale \cite{a115}.

\textbf{2.Data Availability and Quality:} 

Incomplete and Biased Datasets: Protein databases such as UniProt and PDB, although extensive, are biased toward certain organisms (e.g., model organisms such as humans and E. coli) and well-studied proteins. This bias leads to models that perform well on familiar protein families but poorly on underrepresented ones \cite{a111}.

Lack of annotations: Many proteins lack functional annotations or experimental validation, which limits supervised modeling of models for tasks such as function prediction or PPI analysis. The scarcity of high-quality labeled data hampers the ability of models to learn accurate representations \cite{a113}.

Noisy data: Experimental errors, inconsistent labeling, and outdated annotations introduce noise into the datasets, which can mislead the training process and degrade model performance. Cleaning and curating such large datasets is a non-trivial task \cite{a110}.

Limited Structural Data: Although AlphaFold has predicted structures for many proteins,experimentally determined structures remain limited. Structural data is crucial for training models on tasks that require 3D information, such as predicting protein-ligand interactions \cite{a116}.

\textbf{3.Model Interpretability:}

Black-Box Nature: Transformer models are often criticized for being black boxes, providing little insight into how input features contribute to the output predictions. This opacity is problematic in biological contexts where understanding the reasoning behind a prediction is as important as the prediction itself \cite{a117}.

Regulatory Concerns: In clinical applications, regulatory bodies require explanations for decisions made by computational models, especially when they impact patient care. The lack of interpretability in Transformer models poses a barrier to their acceptance in such settings \cite{a118}.

Biological Insight: Without interpretability, it is challenging to derive new biological knowledge from models. Understanding which parts of the protein sequence or structure are most influential in a prediction could lead to discoveries of new functional motifs or interaction sites \cite{a119}.

\textbf{4.Generalization Across Diverse Protein Families:}

Overfitting to Training Data: Transformer models may be overfitting to the protein families represented in the training data, failing to generalize to novel or rare proteins. This limitation reduces their utility in exploring uncharacterized proteins or those from non-model organisms \cite{a113}.

Difficulty with Novel Folds: Proteins with novel folds not seen during training pose a significant challenge. The models may not accurately predict the structures or functions of these proteins due to a lack of prior examples, limiting their applicability in discovering new protein classes \cite{a120}.

Sequence-length variation: Proteins exhibit a wide range of sequence lengths. Transformer models may struggle with extremely long sequences due to computational constraints, or with very short sequences where context is minimal, affecting their performance across the proteome \cite{a110}.

\textbf{5.Multi-modal Integration}

Complexity of biological data: Incorporating diverse data types, such as genomic, transcriptomic, proteomic, and metabolomic data, is inherently complex. Each data type has different characteristics and noise profiles, which complicates the development of models that can effectively integrate them \cite{a121}.

Lack of Unified Frameworks:There is a shortage of modeling frameworks that can seamlessly combine sequence data with structural and functional information. Existing models are often specialized for a single data modality, limiting their ability to capture the full spectrum of protein biology \cite{a122}.

Data Alignment and Synchronization:Aligning data from different sources and ensuring they correspond accurately to the same proteins or biological contexts is challenging. Misalignment can lead to erroneous conclusions and degrade model performance \cite{a123}.

\section{FUTURE DIRECTIONS}

The application of Transformer models in protein informatics has yielded remarkable advancements; however, several promising research avenues remain to be explored. These directions are essential for overcoming current limitations and unlocking the full potential of Transformers in biological and biomedical applications.

\textbf{1.Multi-modal Data Integration}

  One of the most promising directions is the integration of multimodal data into transformer models to enhance the accuracy of quaternary structure prediction. By combining data from various biological sources, including sequence information, structural data, evolutionary patterns, and biochemical properties, models could capture a more comprehensive picture of protein-protein interactions and complex assembly. Specifically:

   1). Structural and Functional Annotations: Incorporating data from cryo-electron microscopy (cryo-EM), X-ray crystallography, and NMR spectroscopy could help Transformer models learn spatial relationships within complexes with high precision. Recent advancements in structural databases, such as AlphaFold’s structural proteome predictions, offer vast amounts of annotated 3D structure data that could be leveraged to train Transformers with 3D spatial awareness, improving quaternary structure predictions significantly \cite{a72}.

  2). Evolutionary and Co-evolutionary Data: Protein complexes often involve residues that have evolved to co-interact. By integrating evolutionary data from multiple sequence alignments (MSAs) with co-evolutionary metrics, Transformers can learn to detect conserved residues critical to protein interfaces. Models like ESM and AlphaFold have already begun using evolutionary signals in structure prediction; expanding this to complex-level predictions could help detect intricate patterns that dictate multimeric assembly \cite{a73} \cite{a74}.

  3). Biophysical Properties: Including data on biophysical properties, such as binding affinity, hydrophobicity, and electrostatic interactions, could allow models to differentiate between transient and stable complexes. Such detailed information would enable Transformers to predict not only whether two proteins interact but also the strength and nature of these interactions, which is crucial for understanding complex stability and functionality.

\textbf{2.Hybrid Modeling Approaches}

  Hybrid modeling approaches, combining deep learning with classical structural biology techniques, offer another compelling direction for quaternary structure prediction. Hybrid methods can provide more accurate and computationally efficient predictions by leveraging the strengths of different modeling techniques:

  1). Combining Transformers with Molecular Dynamics (MD) Simulations: While Transformer models are proficient at predicting static quaternary structures, MD simulations capture the dynamic behavior of protein interactions. A hybrid approach where Transformers predict initial configurations, followed by refinement through MD, could yield both accurate and biologically relevant representations of protein complexes in motion \cite{a75} \cite{a76}.

  2). Integrating Deep Learning with Physics-Based Models: Physics-based methods, such as molecular mechanics, are highly accurate for modeling atomic interactions within protein complexes. Combining these methods with Transformer models would allow initial coarse-grained predictions that can be refined with physics-based details, thus balancing computational efficiency with high-resolution accuracy. This approach could be particularly useful for large complexes, such as viral capsids, where fine-grained modeling is computationally prohibitive.

 3). Leveraging Experimental Data for Model Validation: Incorporating experimentally derived data, such as distance constraints from FRET or cross-linking mass spectrometry, can validate Transformer predictions and correct model outputs based on real-world data. By grounding Transformer predictions in experimental results, hybrid approaches could achieve high confidence and reliability in complex structure predictions.

 \begin{table*}[!t]
\centering
\caption{Published Algorithms}
\label{tab:published-algorithms}
\begin{tabularx}{\textwidth}{ll>{\centering\arraybackslash}X}
\toprule
\textbf{Model} & \textbf{Time} & \textbf{Repository URL} \\
\midrule
AlphaFold & 2021 & \url{https://github.com/deepmind/alphafold} \\
RoseTTAFold & 2021 & \url{https://github.com/RosettaCommons/RoseTTAFold} \\
ESM-Fold & 2022 & \url{https://github.com/facebookresearch/esm} \\
OmegaFold & 2022 & \url{https://github.com/HeliXonProtein/OmegaFold} \\
ProtGPT2 & 2022 & \url{https://huggingface.co/docs/transformers/main_classes/trainer} \\
RFDiffusion & 2023 & \url{https://github.com/RosettaCommons/RFdiffusion} \\
ProteinBERT & 2022 & \url{https://github.com/nadavbra/protein_bert} \\
ProtTrans & 2021 & \url{https://github.com/agemagician/ProtTrans} \\
DeepPPI & 2017 & \url{http://ailab.ahu.edu.cn:8087/DeepPPI/index.html} \\
TAPE & 2019 & \url{https://github.com/songlab-cal/tape} \\
ChemBERTa & 2020 & \url{https://github.com/seyonechithrananda/bert-loves-chemistry} \\
MolBERT & 2021 & \url{https://github.com/cxfjiang/MolBERT} \\
TransConv & 2025 & \url{https://github.com/sayantanDs/transconv} \\
\bottomrule
\end{tabularx}
\end{table*}

\begin{table*}[!t]
\centering
\caption{Summary of Datasets Available for Transformer-Based Protein Research Applications}
\label{tab:published-algorithms1}
\resizebox{\textwidth}{!}{
\begin{tabularx}{\textwidth}{@{} l l l X @{}}
\toprule
\textbf{Model} & \textbf{Dataset} & \textbf{Size} & \textbf{Access Method} \\ 
\midrule
ESM-Fold & UR50 (sample UR90) & 30,051 & \url{https://huggingface.co/datasets/nferruz/dataset_fastas} \\
ProtGPT2 & UR50\_2021\_04 & 9,935,212 & \url{https://huggingface.co/datasets/nferruz/UR50_2021_04.} \\
RFDiffusion & RoseTTAFold Diffusion & 11,714 & \url{https://figshare.com/s/439fdd59488215753bc3} \\
ProteinBERT & UniRef90 & 43.85 GB (compressed) & \url{ftp://ftp.uniprot.org/pub/databases/uniprot/uniref/uniref90/uniref90.fasta.gz} \\
ProtTrans & UniRef100, UniRef50, BFD & 2.3B sequences & \url{http://doi.ieeecomputersociety.org/10.1109/TPAMI.2021.3095381} \\
TAPE & Pfam & 7 GB (compressed) & \url{https://github.com/songlab-cal/tape} \\
ChemBERTa & PubChem-10M & 10,000,000 SMILES & \url{https://pubchem.ncbi.nlm.nih.gov/} \\
MolBERT & ZINC15, ChEMBL27 & 4,000,000 SMILES & \url{https://zinc15.docking.org} \\
AlphaFold & PDB & 170,000 & \url{https://www.rcsb.org/} \\
TransConv & NetSurf, ProteinNet & 11,000 & \url{https://github.com/sayantanDs/transconv/tree/main/attsc_dataset} \\
\bottomrule
\end{tabularx}
}
\end{table*}

\textbf{3. Model Interpretability and Explainable AI}

1). Attention Mechanisms for Model Interpretability:One potential way to improve interpretability is by utilizing the attention mechanism inherent in Transformer models. By visualizing the attention weights, researchers can gain insights into which parts of the protein sequence or structure the model is focusing on when making predictions. These attention maps can help explain why a model predicts a certain protein interaction, function, or structure. For example, attention could focus on regions in the protein sequence that are critical for binding interactions or areas of structural importance \cite{a148}.

2). Explainable AI Techniques:Methods such as SHAP (SHapley Additive exPlanations) and LIME (Local Interpretable Model-agnostic Explanations) can be applied to interpret model predictions, offering a way to understand feature contributions \cite{a130}.

3). Incorporating Domain Knowledge: Embedding biological knowledge into models, such as known functional domains or structural constraints, can enhance interpretability and guide the learning process. Hybrid models that combine machine learning with mechanistic understanding are promising in this regard \cite{a131}.

\textbf{4. Expansion of Training Data and Benchmarks}

Comprehensive and diverse datasets are foundational to model performance. Future work should focus on expanding protein complex databases with more multimeric structures, transient interactions, and annotated functional states. Emphasis should be placed on underrepresented protein families, such as membrane-associated or signaling proteins. The development of standardized benchmarks and evaluation protocols for quaternary structure prediction would also enable more rigorous comparisons across models and facilitate reproducibility \cite{a105}.

\textbf{5. Efficient and Scalable Model Architectures}

Although Transformer models have shown remarkable success in protein informatics, their computational cost remains a significant challenge. Training large-scale Transformer models typically requires vast amounts of computational resources, including powerful GPUs and distributed computing systems, making them inaccessible to many researchers. Reducing computational costs without sacrificing performance is a priority for future research.

1). Sparse Attention Mechanisms: To reduce computational complexity, researchers are exploring sparse attention mechanisms that limit the number of interactions modeled between tokens. Models like Longformer and Performer introduce approximations to the standard attention mechanism, enabling the handling of longer sequences with reduced computational load \cite{a124}.

2). Recurrent Neural Networks and Memory Mechanisms:Incorporating recurrence or memory mechanisms into Transformer architectures could help capture long-range dependencies without processing the entire sequence simultaneously, thus saving computational resources \cite{a125}.

3). Knowledge Distillation: Techniques like knowledge distillation can be used to transfer knowledge from large, cumbersome models to smaller, more efficient ones without significant loss in performance. This approach can make models more accessible and deployable in resource-limited settings \cite{a126}.

4). Hardware Acceleration: Advances in hardware, such as the development of specialized AI chips and neuromorphic computing, could alleviate computational constraints. Utilizing cloud-based resources and distributed computing can also make high-performance modeling more accessible \cite{a127}.

\textbf{6. Interdisciplinary and Collaborative Research}

The complex nature of protein systems necessitates cross-disciplinary collaboration among computational biologists, structural biologists, machine learning experts, and domain scientists\cite{a37}\cite{a54}. Such collaboration will accelerate the translation of Transformer models into practical workflows for drug discovery, synthetic biology, and personalized medicine. Furthermore, co-development of models and wet-lab validation pipelines will ensure that computational predictions have real-world biological relevance.

\section{CONCLUSION}

Transformer models have rapidly become a cornerstone in protein informatics, offering robust mechanisms for learning long-range dependencies and capturing complex biological patterns. This survey has presented a systematic review of Transformer-based architectures and their applications across critical domains, including protein structure prediction, function annotation, protein-protein interactions, and computational drug discovery. Starting from foundational principles, we examined leading models such as AlphaFold, ESM, ProtTrans, and RoseTTAFold, highlighting their design innovations and contributions to protein science.

Beyond summarizing current advances, this work also identified key limitations associated with Transformer models, particularly in terms of computational scalability, data requirements, and biological interpretability. We emphasized the importance of future directions such as multimodal data integration, hybrid modeling with experimental constraints, improved model efficiency, and enhanced domain generalization. Moreover, we advocated for interdisciplinary collaboration as a driver for methodological breakthroughs and real-world applicability.

By focusing specifically on protein-related tasks, this survey provides a unique perspective on the intersection of deep learning and molecular biology. It is our hope that this work will guide future research in developing more efficient, interpretable, and biologically grounded Transformer models, ultimately advancing the frontiers of bioinformatics, structural biology, and personalized medicine.

\section{Resources and Reproducibility}

\subsection{Open Source Implementation}
Open-source implementations of Transformer-based models in protein informatics have played a crucial role in advancing the field, enabling researchers to conduct baseline experiments, benchmark new methods, and foster reproducibility. Publicly available codebases not only facilitate rapid experimentation but also promote transparency and collaboration across the community. TABLE~\ref{tab:published-algorithms} summarizes representative open-source implementations related to protein Transformer models, detailing their publication year and repository links.

\subsection{Datasets Used}
To support the development and evaluation of Transformer models in protein informatics, researchers have employed a diverse range of large-scale biological datasets. These datasets cover protein sequences, structures, molecular properties, and chemical compounds, offering essential training material for pretraining, fine-tuning, and benchmarking. The use of well-curated and standardized datasets has greatly enhanced model generalizability and robustness in real-world bioinformatics scenarios.

Most datasets fall into one of the following categories: (1) large-scale protein sequence corpora such as UniRef, UR50, and BFD; (2) structure-related datasets from PDB or derived sources used for modeling or generative design; and (3) chemical and molecular datasets like SMILES strings for tasks involving drug discovery.

For example, models such as ProtTrans and ProteinBERT were pretrained on billions of protein sequences from UniRef50, UniRef100, and BFD, enabling them to capture both local motifs and global evolutionary patterns. ESM-Fold and ProtGPT2 utilized subsets like UR50 for efficient pretraining. Structure-based generative models such as RFDiffusion further incorporated manually curated datasets for training backbone diffusion networks. In the field of chemical modeling, ChemBERTa and MolBERT used millions of molecular SMILES representations from PubChem and ZINC databases to enable protein-ligand interaction prediction.

TABLE~\ref{tab:published-algorithms1} summarizes the major datasets adopted in the literature, including the dataset name, size, and access method.

\bibliographystyle{IEEEtran}
%\bibliography{references}

\end{document}